\PassOptionsToPackage{table,svgnames}{xcolor}
\documentclass{article}

\usepackage{PRIMEarxiv}

\usepackage[utf8]{inputenc} 
\usepackage[T1]{fontenc}    
\usepackage{hyperref}       
\usepackage{url}            
\usepackage{booktabs}       
\usepackage{amsfonts}       
\usepackage{nicefrac}       
\usepackage{microtype}      
\usepackage{lipsum}
\usepackage{fancyhdr}       
\usepackage{graphicx}       
\graphicspath{{media/}}     
\usepackage{xspace}

\usepackage{multirow} 
\usepackage{algorithm}      
\usepackage{algpseudocode}  

\usepackage[most]{tcolorbox}
\usepackage{enumitem}
\usepackage{listings}
\usepackage{subcaption}
\usepackage{array}
\usepackage{makecell}  
\usepackage[justification=centering]{caption}
\newcommand{\huggingface}{\raisebox{-1.5pt}{\includegraphics[height=1.05em]{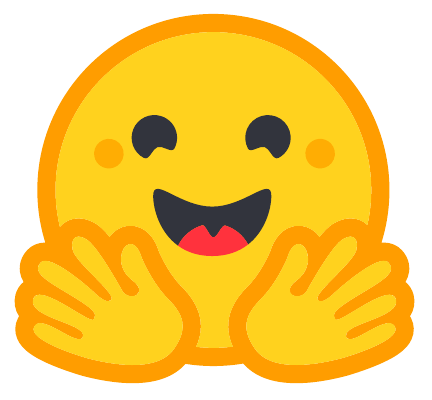}}\xspace}
\newcommand{\github}{\raisebox{-1.5pt}{\includegraphics[height=1.05em]{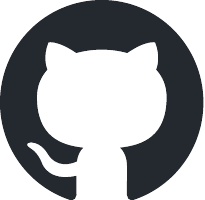}}\xspace}

\title{SciSage: A Multi-Agent Framework for High-Quality Scientific Survey Generation}

\author{
  Xiaofeng Shi\textsuperscript{1}\thanks{\ \ Equal contribution.}%
               \,\,\,\thanks{\ \ Corresponding author. Email: \texttt{xfshi@baai.ac.cn}},
  Qian Kou\textsuperscript{1}\footnotemark[1], 
  Yuduo Li\textsuperscript{1,2}\thanks{\ \ Work done during internship at BAAI.}, 
  Ning Tang\textsuperscript{1,3}\footnotemark[3],
  Jinxin Xie\textsuperscript{1}, 
  Longbin Yu\textsuperscript{1}, 
  Songjing Wang\textsuperscript{1}, 
  Hua Zhou\textsuperscript{1}\thanks{\ \ Project leader.} \\
  \textsuperscript{1}Beijing Academy of Artificial Intelligence (BAAI) \\
  \textsuperscript{2}Beijing Jiaotong University (BJTU) \quad
  \textsuperscript{3}Fudan University (FDU)
}

\begin{document}

\maketitle

\begin{abstract}

The rapid growth of scientific literature demands robust tools for automated survey-generation. However, current large language model (LLM)-based methods often lack in-depth analysis, structural coherence, and reliable citations. To address these limitations, we introduce SciSage, a multi-agent framework employing a reflect-when-you-write paradigm. SciSage features a hierarchical Reflector agent that critically evaluates drafts at outline, section, and document levels, collaborating with specialized agents for query interpretation, content retrieval, and refinement. We also release SurveyScope, a rigorously curated benchmark of 46 high-impact papers (2020–2025) across 11 computer science domains, with strict recency and citation-based quality controls. Evaluations demonstrate that SciSage outperforms state-of-the-art baselines (LLM$\times$MapReduce-V2, AutoSurvey), achieving +1.73 points in document coherence and +32\% in citation F1 scores. Human evaluations reveal mixed outcomes (3 wins vs. 7 losses against human-written surveys), but highlight SciSage’s strengths in topical breadth and retrieval efficiency. Overall, SciSage offers a promising foundation for research-assistive writing tools.

\begin{center}
\begin{tabular}{crl}
 \textbf{Github} & \github & \href{https://github.com/FlagOpen/SciSage.git}{\path{github.com/FlagOpen/SciSage}}\\
 \textbf{Benchmark} & \huggingface & \href{https://huggingface.co/datasets/BAAI/SurveyScope}{\path{BAAI/SurveyScope}}
\end{tabular}
\end{center}

\end{abstract}


\section{Introduction}
The rapid growth of scientific literature, particularly in fast-evolving domains like artificial intelligence, poses increasing challenges for researchers to stay up-to-date\cite{bornmann2021growth,wang2024autosurvey}. As literature accumulation outpaces human synthesis capacity, concerns emerge around research quality, redundancy, and accessibility. Survey articles help address this burden by systematically synthesizing existing work, highlighting key trends, and identifying open problems. High-quality surveys provide structured overviews, critically evaluate methodologies, and guide future research\cite{wang2022deep,hadi2023survey,wang2024survey}. However, their creation remains labor-intensive, demanding deep domain expertise, thematic abstraction, and rigorous citation management. As the scale and speed of academic papers continue to grow, scalable and robust survey generation methods have become increasingly essential.

With the development  of large language models (LLMs)\cite{hurst2024gpt,team2023gemini,guo2025deepseek,qwen3}, researchers are employing them to automate scientific research survey writing. Most prior systems for automating literature surveys adhere to a two-stage pipeline—outline generation followed by content synthesis. AutoSurvey\cite{wang2024autosurvey} employs a streamlined pipeline of retrieval, outline drafting, subsection generation, and evaluation to produce  human-level surveys. STORM\cite{shao2024assisting} leverages multi-agent to generate Wikipedia-style drafts, while Co‑STORM\cite{jiang2024into} adds human-in-the-loop semantic mind-mapping to improve outline coherence. For 
large-scale survey tasks, LLM$\times$MapReduce-V2\cite{wang2025llm} uses convolutional scaling to synthesize coherent drafts from vast corpora. 

These LLM-based methods above highlight significant progress in automating end-to-end survey generation. However, the outputs they deliver still trail expert‐written surveys on several critical dimensions. First, content quality often suffers from shallow synthesis or speculative claims when source passages are thin or noisy. Second, even when topic coverage is adequate, structural coherence is frequently compromised, with sections drifting in granularity, duplicating ideas, or breaking the logic. Third, generated references can be topically irrelevant, outdated, or hallucinated, leading to low citation relevance and poor scholarly reliability. Finally, existing systems lack a dedicated reflection mechanism capable of critiquing and revising drafts instantly, most of them rely on single-pass generation or lightweight post-editing, leaving deeper logical or factual issues unresolved. Recent work on self-reflection for LLMs\cite{shinn2023reflexion,asai2023self,renze2024self} hints at the promise of iterative refinements which can significantly benefit long-form scholar survey generation.

In order to bridge these gaps, we introduce SciSage(\textbf{Sci}entific \textbf{Sage}), a multi-agent framework that operates a \textit{reflect-when-you-write} paradigm. Central to our design is a Reflector agent that permeates generation phases of the workflow. The Reflector systematically audits outlines, section drafts, and complete manuscripts, dispatching refinement whenever substantive or stylistic deficiencies are detected. Collaborating with 
five other specialized agents—Interpreter, Organizer, Collector, Composer, and Refiner—SciSage starts with outline generation, through section-level retrieval and drafting, to final integration and refinement, resulting in surveys that are both structurally coherent and  high-quality in content.

To evaluate the efficiency of our methods, we release SurveyScope, a benchmark covering 11 research topics with recency and citation-based quality controls, providing a rigorous testbed for survey generation. Experiments on SurveyScope show that SciSage significantly surpasses strong state-of-the-art baselines (LLM$\times$MapReduce-V2, AutoSurvey), achieving +2.8 points in document coherence and +32\% in citation F1 scores. Moreover, Human experts prefer SciSage’s drafts on 7 out of 10 topics, praising broader coverage and tighter narrative flow.

Our main contributions can be summarised in three points:
\begin{itemize}
\item We propose SciSage, a multi-agent system that generates well-organized and high-quality survey in an end-to-end pattern.
\item SciSage includes a iterative reflection mechanism at outline, section, and document scopes, enabling principled self-critique and refinement in survey generation.
\item We introduce SurveyScope benchmark to comprehensively evaluate generated survey in  content quality, structural coherence and reference relevance.
\end{itemize}

\section{Related Works}
\textbf{Scientific survey generation.} The automation of scientific survey generation using Large Language Models (LLMs) has garnered significant attention in recent years. Early approaches primarily relied on retrieval-augmented generation (RAG) techniques to synthesize literature. For instance, OpenScholar\cite{asai2024openscholar} introduced a specialized RAG-based LLM capable of answering scientific queries by identifying relevant passages from a vast corpus of open-access papers, achieving citation accuracy on par with human experts. Despite these advancements, challenges persist in ensuring the structural coherence and depth of generated surveys. AutoSurvey\cite{wang2024autosurvey} proposed a two-stage LLM-based method for survey generation, focusing on logical parallel generation to enhance content quality and citation accuracy. Similarly, SurveyForge\cite{yan2025surveyforge} addressed some of these issues by emphasizing outline heuristics and memory-driven generation, aiming to bridge the quality gap between LLM-generated surveys and those written by humans. InteractiveSurvey\cite{wen2025interactivesurvey} took a different approach by introducing a personalized and interactive survey paper generation system. This system allows users to customize and refine intermediate components continuously during generation, including reference categorization, outline, and survey content, thereby enhancing user engagement and output quality. In the realm of long-form article generation, STORM\cite{shao2024assisting}  presented a writing system that models the pre-writing stage by discovering diverse perspectives, simulating multi-perspective questioning, and curating collected information to create comprehensive outlines, while Co‑STORM \cite{jiang2024into} extends this with human-in-the-loop and semantic mind-map techniques to enhance outline coherence. To handle ultra-long document synthesis, LLM×MapReduce‑V2\cite{wang2025llm} applies entropy-driven convolutional scaling to assemble coherent survey drafts from extensive corpora. In addition, Deep Research tools based on advanced closed-source LLMs\cite{openaiDeepResearch,googleDeepResearch} show promise performance in synthesizing large amounts of online information into comprehensive scitific surveys, whose mechanisms are still unclear. Despite the impressive performance of Deep Research tools, due to closed-source nature, their search mechanisms are still unknown. These systems demonstrate LLMs’ potential in automating end-to-end survey generation, yet persistent challenges remain in guaranteeing content quality, structural comprehensive, and establishing rigorous evaluation standards. 

\textbf{LLM-based Multi-Agent Systems.} 

Recent advancements in LLMs have catalyzed significant progress in multi-agent systems~\cite{li2024survey,guo2024large}, leading to collaboration, specialization, and emergent behavior through structured architectures and dynamic coordination~\cite{hong2023metagpt,wu2023autogen,wang2023unleashing}. Generative Agents~\cite{park2023generative} introduced a framework where 25 AI agents, each with unique identities and memories, autonomously coordinated social events and daily activities in a virtual town, demonstrating believable human-like behavior. MetaGPT~\cite{hong2023metagpt} utilizes human-like standard operating procedures and specialized roles—product manager, architect, coder, tester—to reduce hallucinations in software generation. AutoGen~\cite{wu2023autogen} offers a flexible framework allowing configurable conversation patterns among LLM agents, enabling tool invocation, human-in-the-loop interventions, and multi-agent debate strategies to boost reasoning and factuality. AgentVerse~\cite{chen2023agentverse} emphasizes group formation and emergent social behaviors, demonstrating performance gains from collaborative diversity. ChatDev~\cite{qian2023chatdev} and its companion systems implement entire virtual development teams, validating structured role allocation on real‑world code bases. DyLAN\cite{liu2023dynamic} introduced a dynamic LLM‑agent network leveraging inference‑time agent selection via an unsupervised Importance Score, flexible communication structures, and early stopping. Debate oriented frameworks\cite{liang2023encouraging,du2023improving} formalize structured argumentation among solver agents, mediated by aggregators, and show clear improvements on arithmetic and reasoning benchmarks.  AgentNet\cite{yang2025agentnet} introduced a decentralized, retrieval‑augmented, evolutionary coordination model over a dynamically evolving DAG network, eliminating central orchestrators and enabling scalable, privacy-aware specialization. We share the idea by establishing an LLM-based multi-agent system to facilitate academic research.

\label{sec:method}
\begin{figure}[htbp]
    \centering
    \includegraphics[width=1\linewidth]{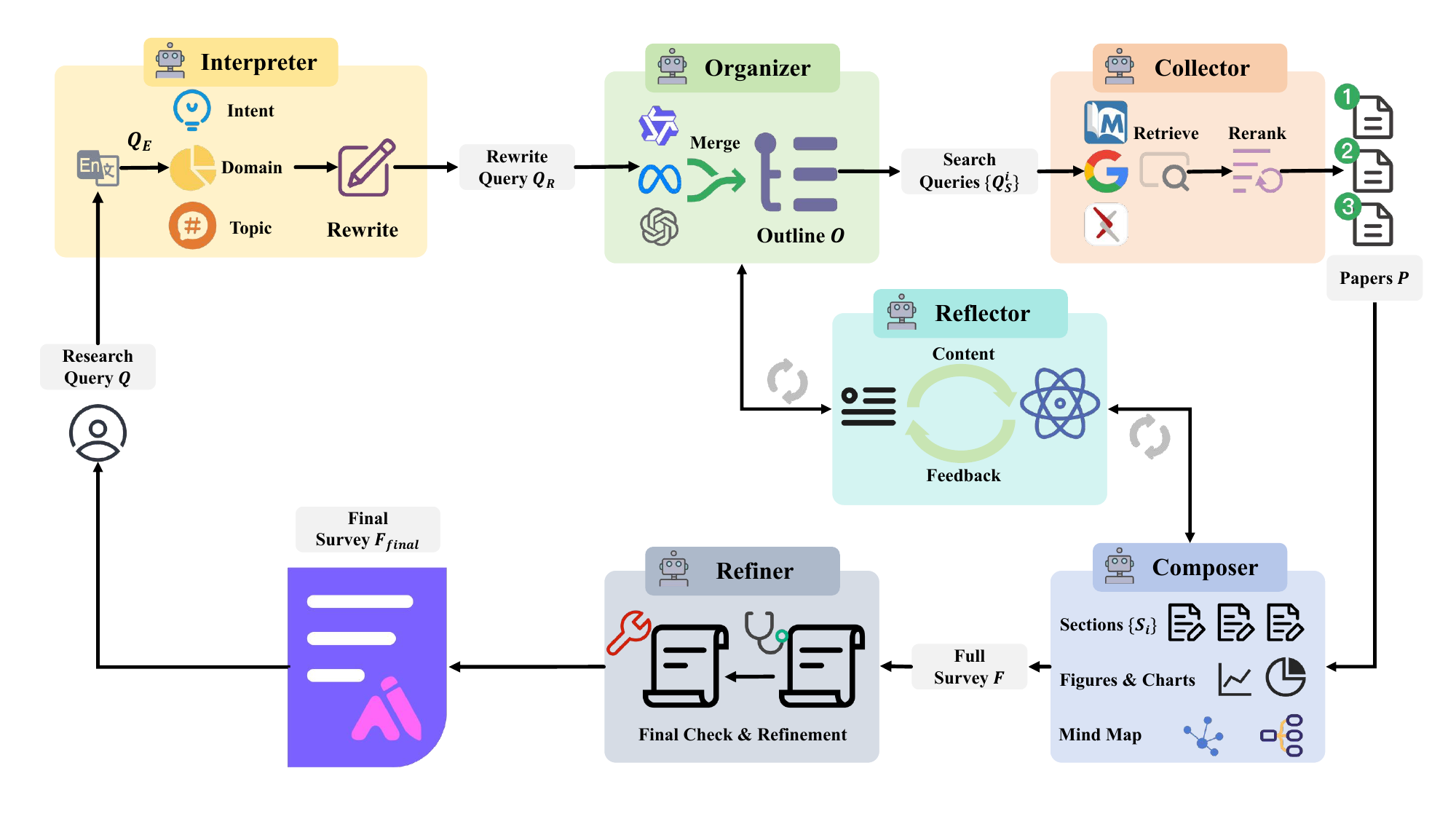}
    \caption{Overview of the \textsc{SciSage} framework.}
    \label{fig:overview}
\end{figure}

\section{Method}
In this section, we introduce SciSage, a LLM-based Multi-Agent framework designed for automated scientific survey generation. Inspired by the cognitive and iterative behaviors of expert authors, SciSage leverages a coordinated architecture of specialized agents that unfolds through three interconnected components\textemdash Query Understanding and Rewrite, Retrieval and Content Generation, and Iterative Hierarchical Reflection. Each component comprises distinct agents that cooperate to produce high-quality scientific surveys.

As shown in Figure ~\ref{fig:overview}, SciSage operates as a dynamic, iterative workflow. The system begins with user queries and proceeds through modular stages where intermediate results are critically reviewed and enhanced. Central to this process is the Reflector agent, which simulates expert-like revision cycles to ensure coherence, depth, and informativeness across all sections of the survey. The following subsections provide an in-depth analysis of each module’s architecture and operational logic. The overall pseudo-code of SciSage is summarized in Algorithm ~\ref{alg:scisage}.

\begin{algorithm}[t]
\caption{\textsc{SciSage}: A Multi-Agent Survey Generation System}
\label{alg:scisage}
\begin{algorithmic}[1]
\Require{User query $Q$, research sources $D$, reflection trials $N$,
outline templates $T$}
\Ensure{Final refined survey document $F_{final}$}
\State Rewrite the query and get intent information $Q_R, I \leftarrow \textsc{Interpreter}(Q)$
\State Select a suitable outline template $t \leftarrow \textsc{Organizer}(Q_R, I,T)$
\State Generate outline $O \leftarrow \textsc{Organizer}(Q_R, I, t)$
\Repeat
    \State Receive feedback from Reflector $\Delta O \leftarrow \textsc{Reflector}(O,Q_R,I)$
    \If{$\Delta O \neq \emptyset$}
        \State Refine and update $O \leftarrow \textsc{Organizer}(O,\Delta O,t)$
    \EndIf
\Until{$\Delta O = \emptyset$ \textbf{or} max reflection trails $N$ reached}

\State Construct search queries for each section in final outline $\{Q_S^i\}_{i=1}^K \leftarrow \textsc{Organizer}(O)$
\ForAll{$\text{outlined section } s_i \in O$}
    \State Retrieve relevant papers from multiple sources ${P}_{i} \leftarrow \textsc{Collector}(S_i, {D})$
    \State Generate section content $S_{i} \leftarrow \textsc{Composer}(s_i,{P}_{i})$
    \Repeat
        \State Receive feedback from Reflector $\Delta S_i \leftarrow \textsc{Reflector}(S_i,P_i,O)$
        \If{$\Delta S_i \neq \emptyset$}
            \State Refine and update $S_i \leftarrow \textsc{Composer}(S_i,{P}_{i}, \Delta S_i)$
        \EndIf
    \Until{$\Delta C_i = \emptyset$ \textbf{or} max reflection trails $N$ reached}
\EndFor
\State Integrate all sections to full survey $F \leftarrow \text{Merge}(S_{1},\dots,S_{K})$
\State Refine and get the final survey $F_{final} \leftarrow \textsc{Refiner}(F)$
\State \Return $F_{final}$
\end{algorithmic}
\end{algorithm}

\subsection{Query Understanding and Rewrite}
The efficacy of the entire review generation process is contingent upon a precise and comprehensive understanding of the user's request. The Interpreter Agent serves as the entry point of the SciSage framework. Its objective is to transform original, often ambiguous user queries into well-structured, standardized, and actionable instructions for downstream agents. First, to accommodate multilingual user queries, the Interpreter performs automatic language detection for query $Q$ and translates it into English $Q_E$. This standardization ensures consistency in downstream processing and leverages broader retrieval sources.
Next, the Interpreter engages in a deep semantic analysis of the translated query to discern the user's core intent, scientific domain of interest, and research topic, which can be represented as intent information $I$. For example, given the query \textit{"The latest progress in code generation using LLM"}, the Interpreter infers that the user seeks \textit{recent advances} in \textit{deep learning} of \textit{LLMs for code generation}.
Finally, to maximize the precision and recall of the information retrieval phase, the initial query often requires refinement. The Interpreter evaluates whether the input query needs to be rewritten. Once ambiguity, vagueness, or informal phrasing is detected, the Interpreter generates a refined version $Q_R=\text{Interpreter}(Q_E, I)$ that is semantically equivalent but structurally optimized for retrieval and generation purposes. Prompts for query understading and rewriting are shown in Appendix ~\ref{appendix:prompt for query understanding}.

\subsection{Retrieval and Content Generation}
The central engine of the SciSage framework executes a "bottom-up" workflow for content creation, moving from high-level planning to detailed writing and final polishing. 
This entire process is orchestrated across four specialized agents\textemdash Organizer, Collector, Composer, and Refiner—working in unison to produce a coherent and comprehensive survey.

\textbf{Outline Construction} The Organizer Agent constructs a comprehensive, logical, and scholarly outline that reflects the user’s intent, guiding high-quality content generation. It begins by selecting a suitable outline structure from a curated template library $T$ based on the user's intent(e.g., survey, theory) detected by the Interpreter. To move beyond this initial template and foster a more innovative and robust structure,  the Organizer then employs a multi-model ensemble strategy. Multiple LLMs generate varied outline candidates in parallel to promote 
diversity and reduce bias. These candidate outlines are synthesized into a unified structure using content-aware heuristics and the outline is represented as $O=\textsc{Organizer}(Q_R,I,t)=\text{Merge}(O_{LLM_1},\cdots,O_{LLM_N})$. Finally, for each section and subsection in the outline, the Organizer extracts key points and generates precise search queries $\{Q_S^i\}$ to guide the following retrieval process, while deliberately excluding non-content sections such as conclusion and references.
The ultimate output of this stage is a tree structure where each node contains a section title, its hierarchy, key points, and the corresponding search queries, which is then passed to the Collector to initiate the research phase.

\textbf{Multi-Source Retrieval and Re-Ranking} The Collector Agent serves as the research assistant and gathers high-quality references from various academic sources. Integrated with APIs from multiple scholarly sources $D$ (e.g., arXiv, PubMed, Google Scholar), the Collector employs a multi-source adaptive retrieval strategy. Guided by the domain context provided by the Interpreter, it prioritizes sources most likely to yield relevant results, thereby improving both the efficiency and precision of the retrieval process.
Once the relevant sources are identified, the Collector retrieves candidate papers and scores them, evaluating their semantic relevance and topical depth. To further ensure the credibility and currency of selected papers, the Collector reranks the retrieval results based on publication recency, venue prestige, author influence, and citation metrics, ultimately prioritizing the most authoritative and timely literature for subsequent content generation. The retrieval process for each section $S_i$ can be represented as ${P}_{i}=\textsc{Collector}(S_i, {D})$, where $P_i$ is the final reranked most relevant paper list.

\textbf{Hierarchical Content Generation} The Composer Agent is the central synthesis engine in the SciSage framework, tasked with transforming the Organizer’s structured outline and the Collector’s curated papers into a coherent and comprehensive scientific survey. It adopts a bottom-up methodology that emphasizes local coherence and factual grounding before scaling up to larger textual structures. The Composer begins with atomic content generation, producing focused, citation-rich content $S_i$ for each outlined subsection $s_i$ by synthesizing titles, abstracts, and full texts from corresponding retrieved papers $P_i$. These atomic units are then assembled into coherent sections $S_{i} = \textsc{Composer}(s_i,{P}_{i})$, each featuring an introductory overview, core discussions and a conclusion. During this process, the Composer also performs key figures and tables extraction and integrates them into section contents, scanning documents (e.g., LaTeX files from arXiv) to heuristically identify and extract visual content that best supports the topic, particularly in method or result sections. Once all sections are generated, the Composer organizes the content at both section and document levels, integrating the components into chapters and compiling them into a full-document draft $F$. This also includes crafting the Introduction and Conclusion/Future Work sections to ensure thematic and logical coherence. To further enhance readability, the Composer generates visual aids such as mindmaps derived from the outline and integrates them with the document to provide a high-level overview of the paper’s structure and intellectual architecture. Mindmap example can be found in Appendix ~\ref{appendix:mindmap-example}.

\textbf{Final Refinement} The Refiner is the final agent in the content generation process, responsible for transforming the draft into a polished document which is ready for publication. Following the Composer's draft generation, a thorough finalization process is conducted by the Refiner for both content and presentation to get the final refined survey  $F_{final} =\textsc{Refiner}(F)$. It improves the internal flow of paragraphs, eliminates redundancy, enforces consistent terminology, and ensures logical transitions throughout the manuscript.  It  starts with the content and citation, where the Refiner progressively aligns the document with the final outline based on the section titles and their 
corresponding content, removes the duplicated references and renumbers the citations. Next, the writing format and style are checked and standardized to meet the academic requirements, while ensuring the clarity of the topic. Lastly, as for the output, the Refiner exports the document in formats such as LaTeX and Markdown to support most publishing systems.

\subsection{Iterative Hierarchical Reflection} The Reflector Agent is a critical innovation of SciSage’s system, functioning as a pervasive, iterative mechanism embedded deeply within the workflows of both the Organizer and Composer. Rather than being a standalone step, it operates through a continuous "generate-reflect-regenerate" loop that drives recursive, multi-level content refinement, mirroring the self-corrective nature of expert academic writing. Its hierarchical scope of reflection spans the entire generation process. At the outline level, the Reflector evaluates outline $O$ in completeness, logical structure, topical relevance, and alignment with academic standards, returning feedback $\Delta O$ to the Organizer for iterative refinement until a quality threshold is reached. At the section level, as the Composer produces section content $S_i$, the Reflector gives critique $\Delta S_i$  in accuracy, evidential support, structural clarity, and the balance of perspectives. If deficiencies are detected, it may trigger new literature retrieval by the Collector, followed by targeted content regeneration. At the full-text level, the Reflector deploys a panel of LLM agents simulating expert personas, such as journal editors, senior professors, and peer reviewers, to evaluate the manuscript from diverse critical views. A majority vote system identifies suboptimal sections, prompting the creation of a structured revision plan, including new key points and queries, which reactivates the Collector and Composer in a recursive improvement cycle. The Reflector also ensures that chapter introductions communicate each chapter’s intent and structure. Through this dynamic process, SciSage transforms initial drafts into rigorously refined academic 
surveys that have withstood multiple rounds of critique and enhancement.

\section{Benchmark}
\label{sec:benchmark}

To comprehensively evaluate the quality of generated survey content, we introduce \textbf{SurveyScope}, a high-quality benchmark specifically designed for academic survey writing. \textbf{SurveyScope} significantly improves upon existing evaluation benchmarks like \textsc{SurveyEval\_Test}~\cite{wang2025llm} and \textsc{AutoSurvey}~\cite{wang2024autosurvey} by enhancing both the diversity of research topics and the quality of papers.

Paper quality in \textbf{SurveyScope} is defined by two key criteria: publication recency and citation count.

Given the fast-moving nature of computer science research—especially in areas such as large language models (LLMs) and AI safety—recent papers are more likely to reflect current trends, methods, and state-of-the-art advances. To ensure timeliness, all papers in \textbf{SurveyScope} were published between 2020 and 2025. The majority are concentrated in the 2023-2024 period, coinciding with the surge in large language model research following the advent of ChatGPT~\cite{roumeliotis2023chatgpt}.

Citation count serves as a proxy for academic influence and recognition. A high citation count generally indicates an influential, well-received, and widely adopted paper. Papers in \textbf{SurveyScope} exhibit significantly higher citation metrics than those in other benchmarks, with a maximum of 2,184 and an average of approximately 322 citations.

These stringent metrics ensure the benchmark's high reliability and representativeness, grounding evaluation results in authoritative and influential literature.

\subsection{Construction Methodology}
\label{sec:bench_create}

The construction pipeline of \textbf{SurveyScope} is illustrated in Figure~\ref{fig:construction_pipeline}, and it consists  the following key steps:

\begin{figure}[ht]
    \centering
    \includegraphics[width=\linewidth]{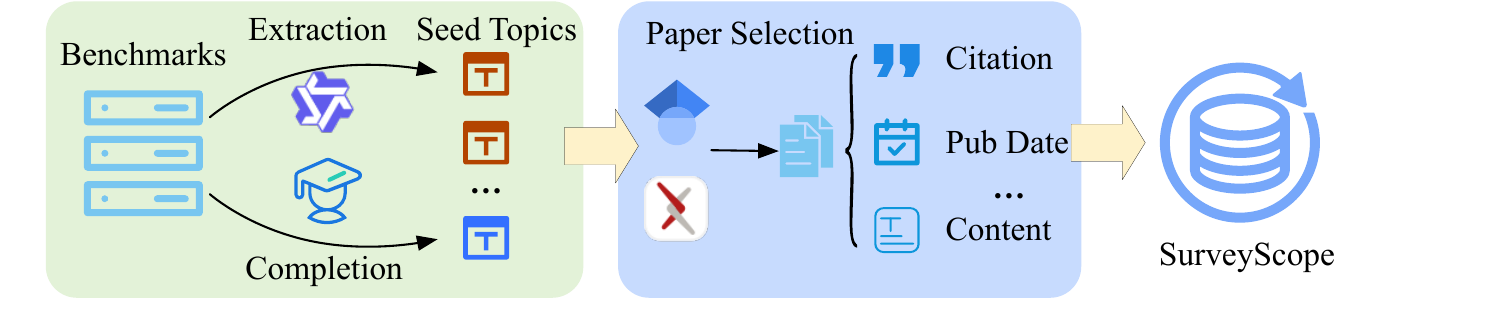}
    \caption{Overview of the \textbf{SurveyScope} construction pipeline.}
    \label{fig:construction_pipeline}
\end{figure}

\begin{enumerate}
    \item \textbf{Domain Extraction from Existing Benchmarks.} We began by collecting open-source benchmark datasets and identifying their covered academic research domains. This extraction was performed using the Qwen3-32B model~\cite{qwen3}, guided by a structured prompt (see Appendix~\ref{appendix:Prompt for Topic Classification}).
    
    \item \textbf{Topic Completion with Expert and LLM Assistance.} To ensure comprehensive coverage, we augmented the initial domain list by incorporating suggestions from both human experts and large language models (LLMs). This step aimed to uncover missing or underrepresented research areas. The prompt used for this stage is detailed in Appendix~\ref{appendix:Prompt for Topic Completion}.
    
    \item \textbf{Paper Selection for Each Domain.} For each identified research domain, we manually searched Google Scholar, prioritizing recent publications with high citation counts to ensure impact and timeliness.
\end{enumerate}

As Table~\ref{tab:benchmark_comparison} shows, the dataset comprises 20 surveys from \textsc{SurveyEval\_Test}, 8 from \textsc{AutoSurvey}, and 18 manually curated surveys collected from Google Scholar and other academic platforms. This diverse sourcing strategy ensures a balanced benchmark that reflects both standardized evaluations and high-quality, real-world survey writing.

\begin{table}[ht]
    \centering
    \small
    \begin{tabular}{ccc}
        \toprule
        \textbf{SurveyEval\_Test} & \textbf{AutoSurvey} & \textbf{Expand Manually Curated} \\
        \midrule
        20 & 8 & 18 \\
        \bottomrule
    \end{tabular}
    \vspace{6pt}   
    \caption{Source distribution of \textbf{SurveyScope}}
    \label{tab:benchmark_comparison}
\end{table}

Following this pipeline, we constructed a curated benchmark of 46 high-quality research papers. Each paper was manually selected by professionals with graduate-level training in computer science, based on criteria including publication recency and citation impact.

\subsection{Characteristics}
\label{sec:bench_character}
\begin{figure}[ht]
    \centering
    \includegraphics[width=0.6\linewidth]{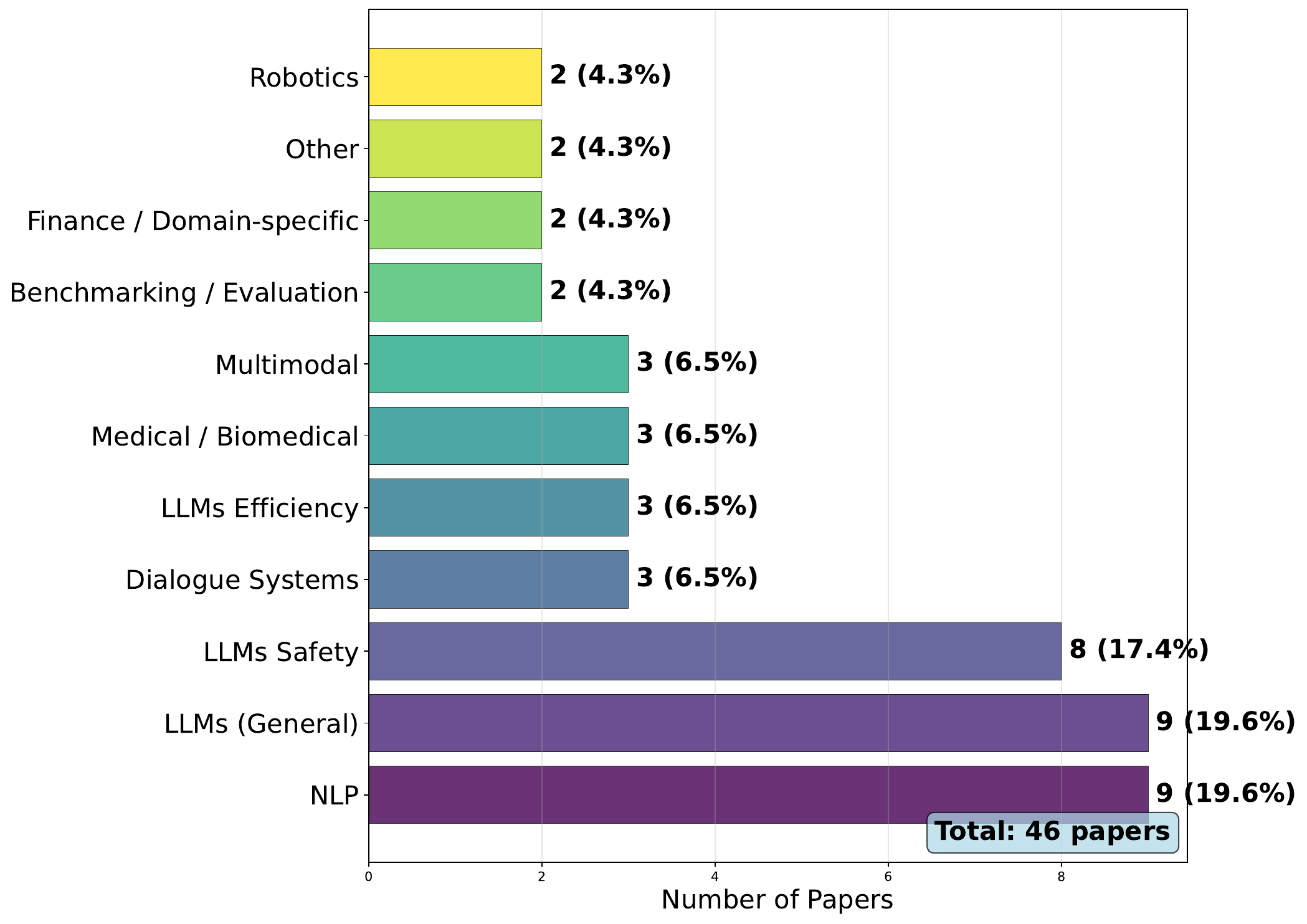}
    \caption{Distribution of topics in \textbf{SurveyScope}}
    \label{fig:topic_distribution_internal}
\end{figure}

Thanks to a carefully designed construction pipeline, \textbf{SurveyScope} exhibits several key characteristics that distinguish it from existing benchmarks:

\paragraph{Broad Topic Coverage} \textbf{SurveyScope} covers a broad range of active research areas in computer science, including natural language processing (NLP), large language models (LLMs), AI safety, robotics, and multimodal learning. This topical diversity enables systematic and cross-domain evaluation of automatic survey generation systems. Figure~\ref{fig:topic_distribution_internal} provides an overview of the topic distribution, with 46 papers spanning 11 distinct topics. A detailed comparison of topic categories across benchmarks is provided in Appendix~\ref{appendix:benchmark-compare-category}.

\begin{figure}[ht]
    \centering
    \begin{subfigure}[b]{0.45\linewidth}
        \centering
        \includegraphics[width=\linewidth]{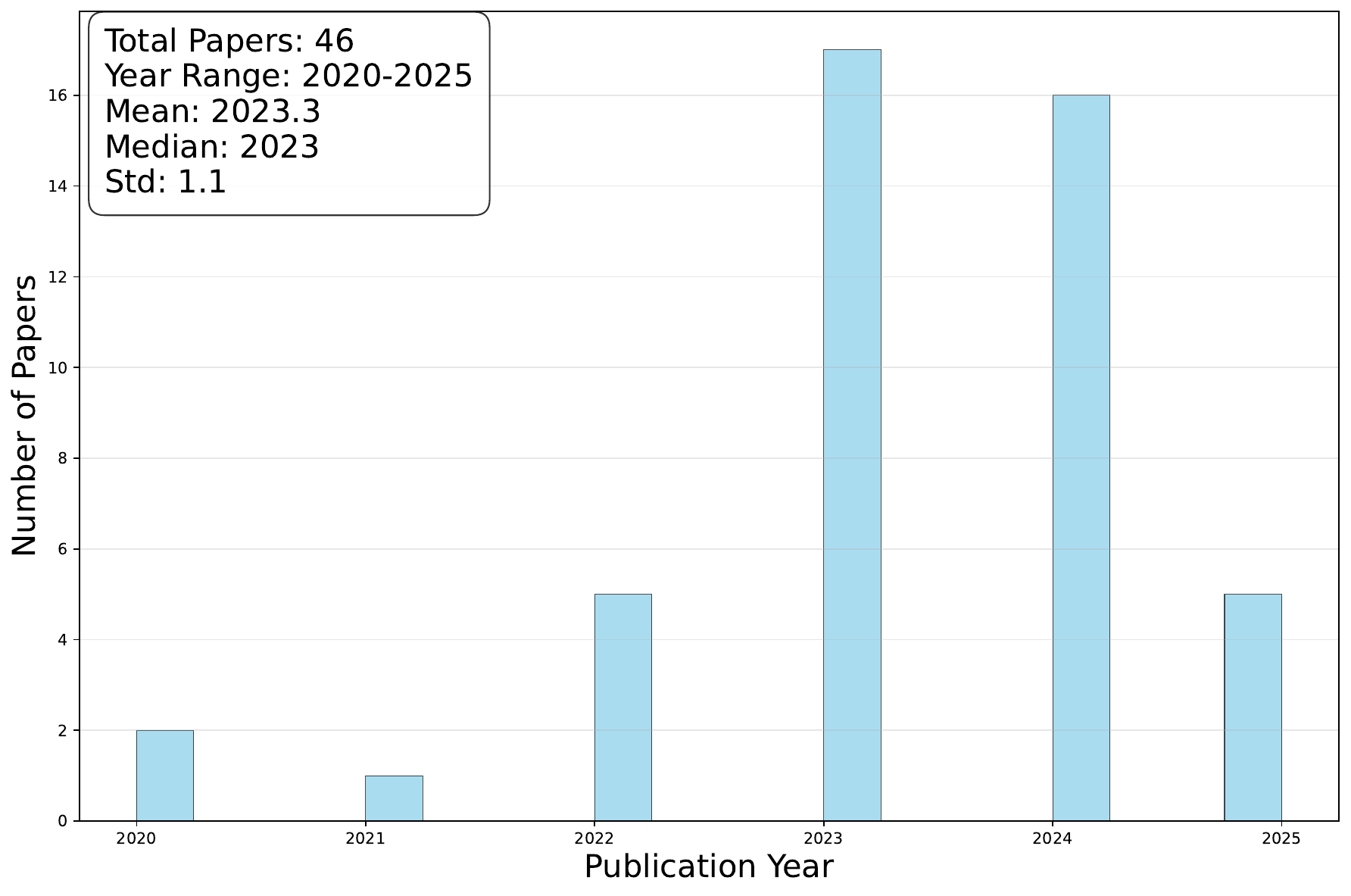}
        \caption*{(a)}
        \label{fig:year_histogram}
    \end{subfigure}
    \hspace{0.05\linewidth}
    \begin{subfigure}[b]{0.45\linewidth}
        \centering
        \includegraphics[width=\linewidth]{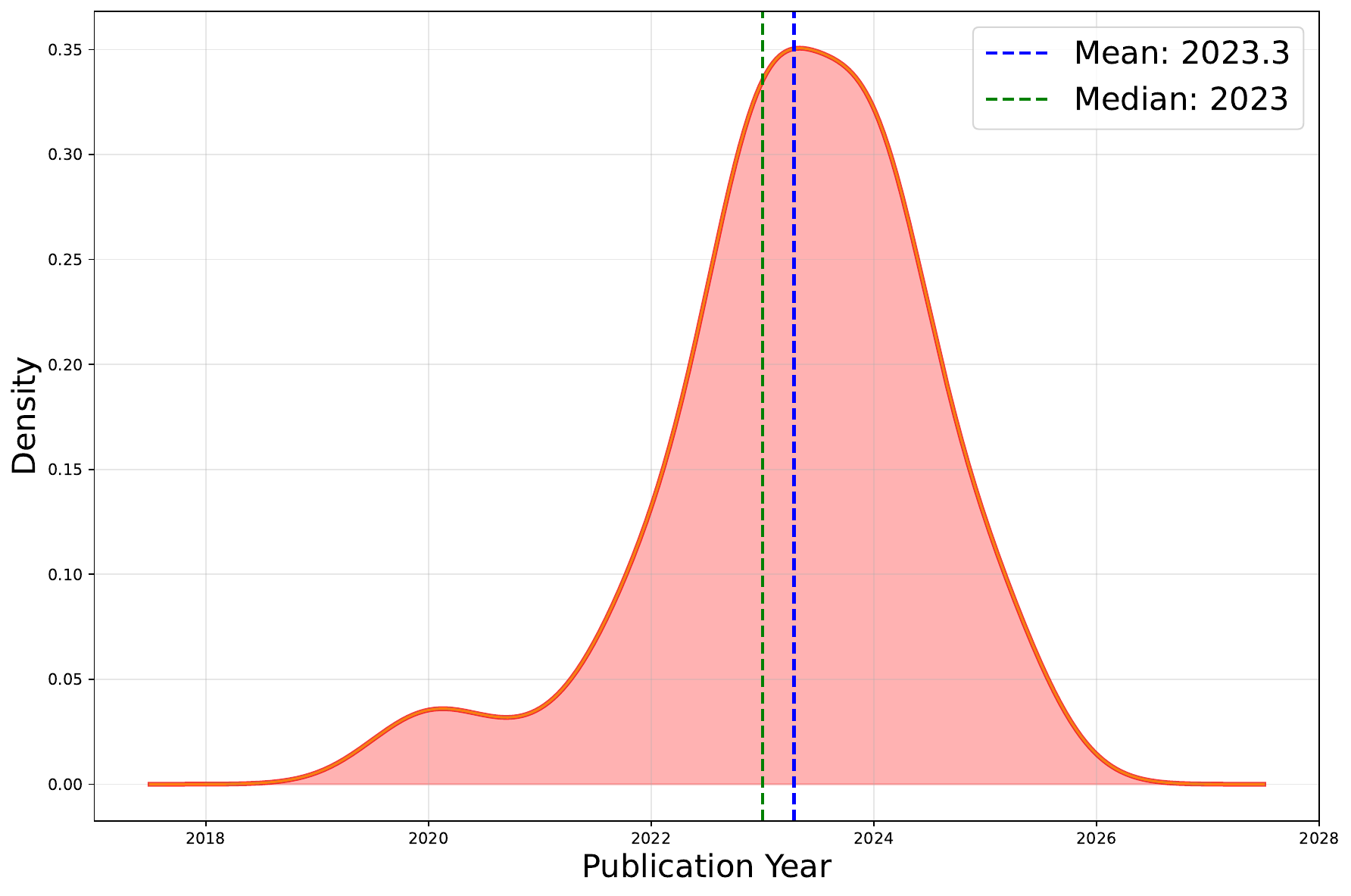}
        \caption*{(b)}
        \label{fig:year_density}
    \end{subfigure}
    \caption{Publication year distribution in \textbf{SurveyScope}. (a) Histogram showing the number of papers published each year. (b) Kernel density estimate illustrating publication trends over time.}
    \label{fig:year_distribution}
\end{figure}

\paragraph{Recency of Publications} \textbf{SurveyScope} includes papers published between 2020 and 2025, capturing recent advances in computer science. The release of ChatGPT ~\cite{roumeliotis2023chatgpt} in late 2022 led to a sharp increase in large language model (LLM) research, resulting in a notable concentration of publications in 2023 and 2024. Figure~\ref{fig:year_distribution} illustrates the temporal distribution of publication years. A comparative analysis of recency across benchmarks is provided in Appendix~\ref{appendix:benchmark-compare-recenty}.

\paragraph{High Citation Count} Papers in \textbf{SurveyScope} demonstrate strong citation impact. The most cited paper received 2,184 citations, and the average is 322 citations per paper. Notably, over 52\% of the papers have been cited more than 100 times, underscoring the benchmark’s focus on influential and widely recognized work. Citation distribution and aggregate statistics across benchmarks are shown in Figure~\ref{fig:citation_comparison}, with further comparisons in Appendix~\ref{appendix:benchmark-compare-citation}.

\begin{figure}[ht]
    \centering
    \begin{subfigure}[b]{0.45\linewidth}
        \centering
        \includegraphics[width=\linewidth]{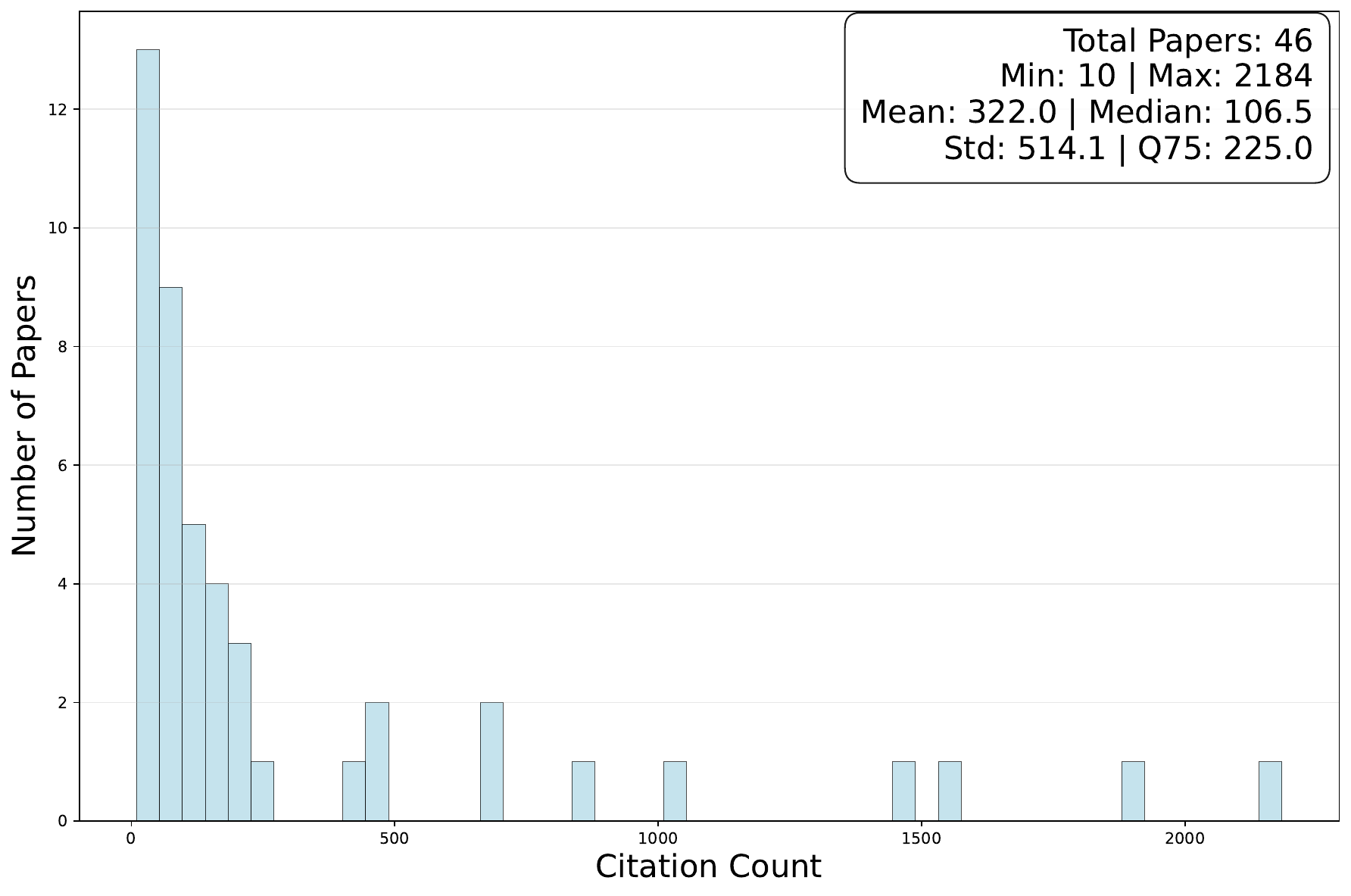}
        \caption*{(a)}
        \label{fig:survey_citation_distribution}
    \end{subfigure}
    \hspace{0.05\linewidth}
    \begin{subfigure}[b]{0.45\linewidth}
        \centering
        \includegraphics[width=\linewidth]{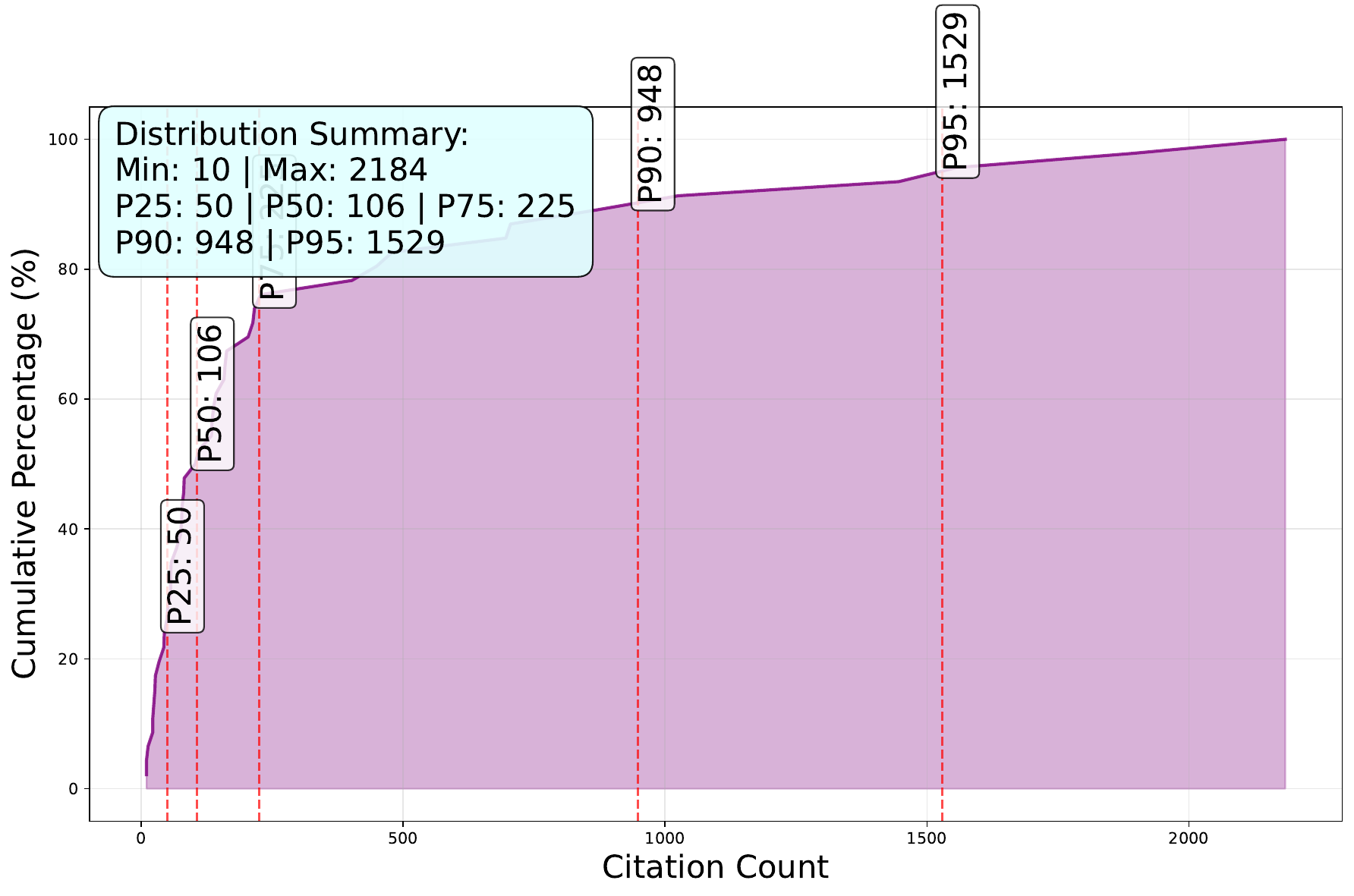}
        \caption*{(b)}
        \label{fig:survey_citation_cumulative}
    \end{subfigure}
    \caption{Citation statistics for \textbf{SurveyScope}. (a) Distribution of citation counts across the dataset. (b) Cumulative plot showing that over 52\% of the papers have received more than 100 citations.}
    \label{fig:citation_comparison}
\end{figure}

\begin{figure}[ht]
    \centering
    \includegraphics[width=0.5\linewidth]{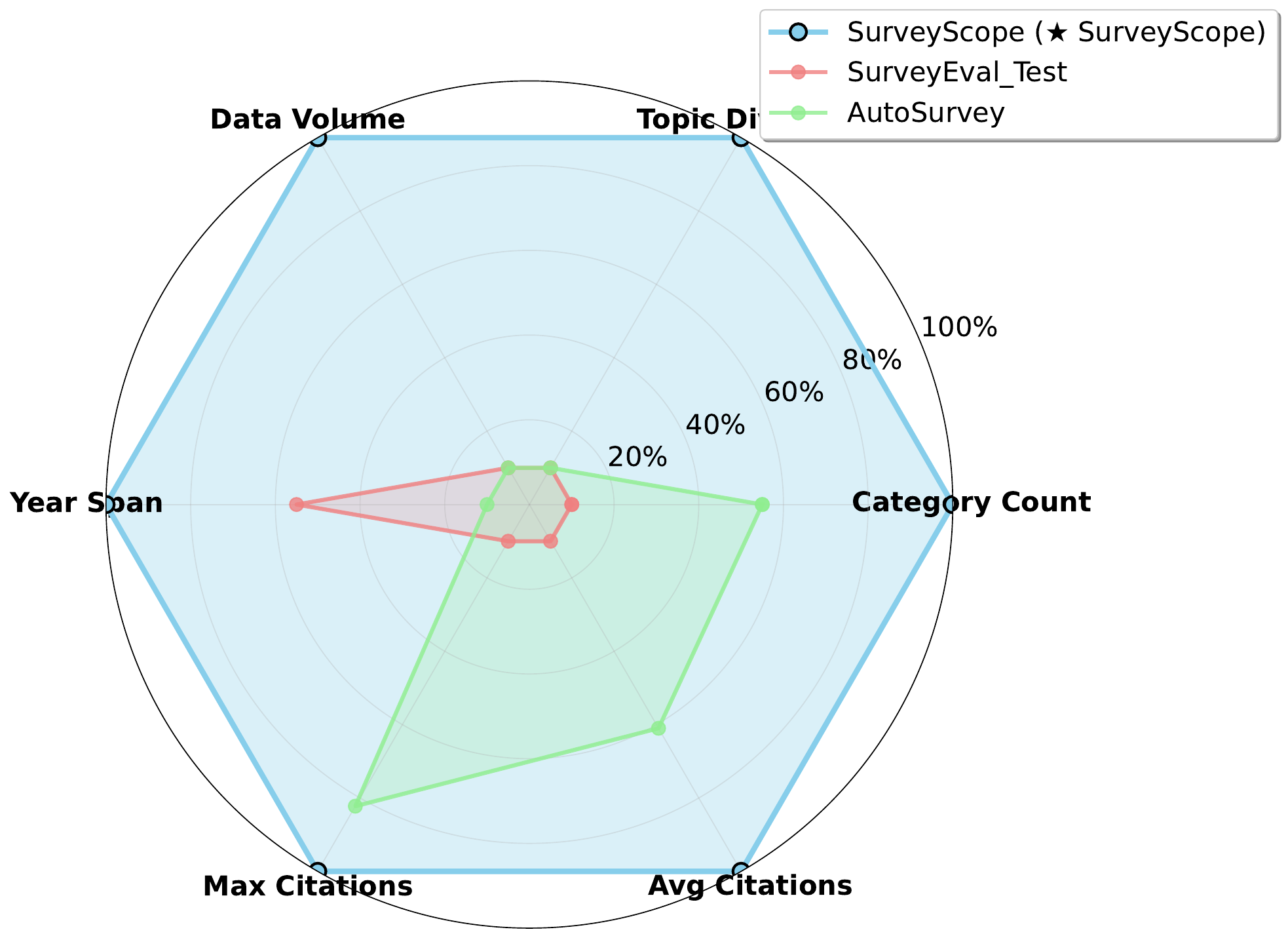}
    \caption{Comparison of different benchmarks on \textit{Category Count}, \textit{Topic Diversity}, \textit{Data Volume}, \textit{Year Span}, \textit{Max Citations}, \textit{Avg Citations}}
    \label{fig:benchmark-compare-pie}
\end{figure}

\paragraph{Summary}
\textbf{SurveyScope} stands out from existing benchmarks through its broad topical coverage, inclusion of recent high-impact publications, and emphasis on citation-based influence. These characteristics make it a comprehensive and reliable resource for evaluating academic survey generation systems. A comparative analysis across benchmarks is presented in Figure~\ref{fig:benchmark-compare-pie}, where \textbf{SurveyScope} consistently leads across all dimensions. Comparison details can be found in Appendix~\ref{appendix:benchmark-compare-details}.

\section{Evaluation}
\label{sec:evaluation}

To comprehensively evaluate the quality of generated content compared to human-written counterparts, we adopt a two-fold evaluation protocol: (1) automatic evaluation leveraging large language models (LLMs), and (2) human evaluation by domain experts.

\subsection{Automatic Evaluation with LLM-based Metrics}

We established an automated evaluation framework, drawing inspiration from \textsc{AutoSurvey}~\cite{wang2024autosurvey} and \textsc{LLM~×~MapReduce-V2}~\cite{wang2025llm}. Our evaluation assesses content across three core dimensions: \textbf{Content Quality}, \textbf{Document Structure}, and \textbf{Reference Accuracy}. All scores are normalized to a 0--100 scale, with higher scores indicating better performance.

\subsubsection{Content Quality Assessment}

\begin{figure}[ht]
    \centering
    \includegraphics[width=0.6\linewidth]{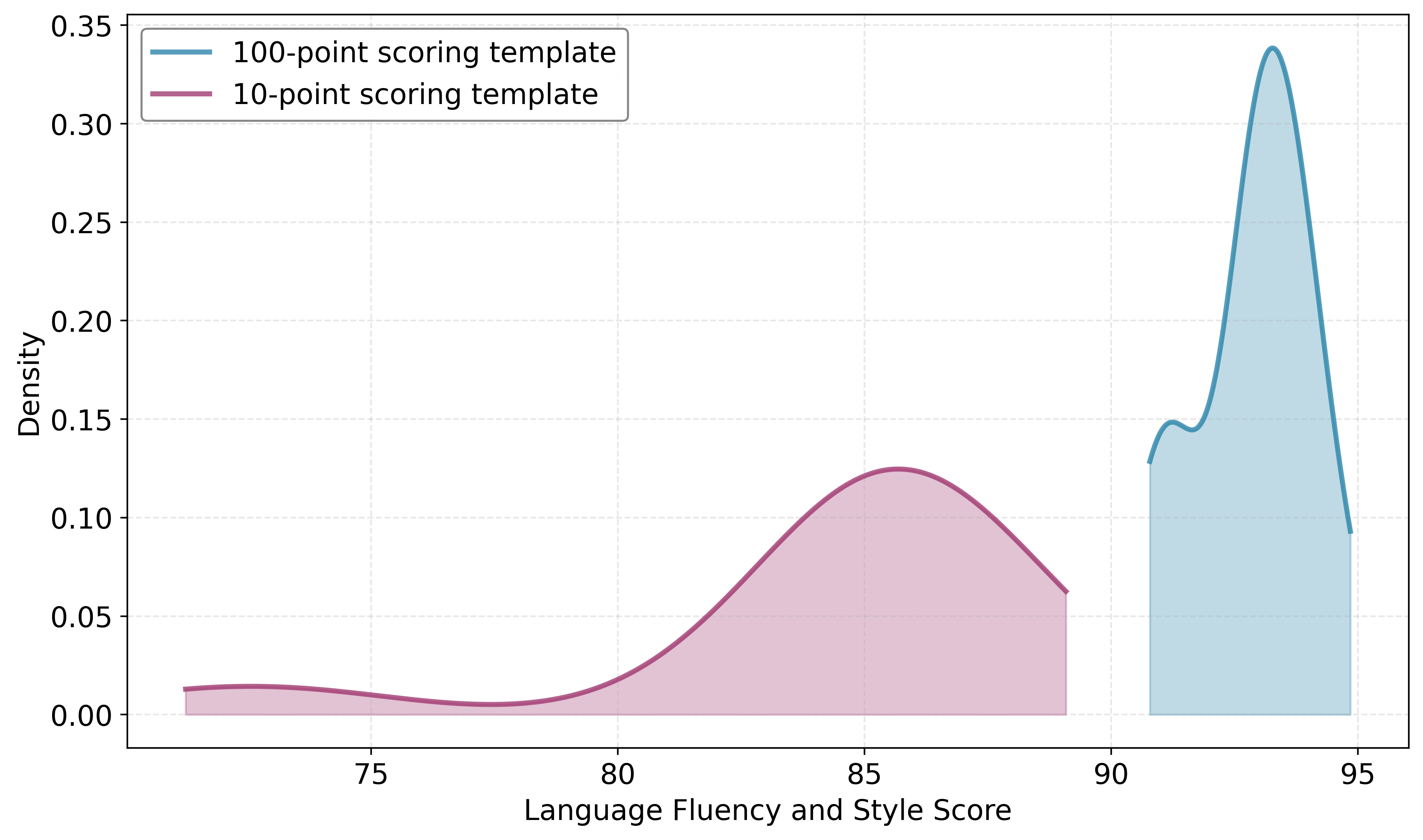}
    \caption{Score comparison: direct 100-point (blue) vs. scaled from 10-point rubric (purple).}
    \label{fig:language-scale}
\end{figure}

We evaluated textual quality across the following dimensions:
\paragraph{Language Fluency and Style} This metric assesses the linguistic quality of generated content, emphasizing academic formality, clarity, and fluency. Referring to the evaluation method provided by \textsc{LLM~×~MapReduce-V2}~\cite{wang2025llm}, we observed that directly using their original 100-point prompt template often resulted in limited score variance, with most outputs receiving uniformly high scores and thus exhibiting low discriminative capacity. To address this, we employed a 10-point scoring rubric to encourage more granular distinctions, then linearly rescaled the scores to a 0--100 range for comparability across evaluation metrics. Figure~\ref{fig:language-scale} presents the score distribution under different prompt templates, demonstrating that the 10-point rubric yields a broader and more informative spread. The prompt details we used can be found in Appendix~\ref{appendix:language-prompt}.

\paragraph{Critical Thinking and Originality} This dimension evaluates the depth of analysis, the originality of perspectives, and the articulation of forward-looking insights. To ensure consistency and interpretability, we designed structured prompts that elicit both numerical ratings and textual justifications from the model. Following the observation that the 100-point scale used in \textsc{LLM~×MapReduce-V2}\cite{wang2025llm} often results in compressed score distributions, we adopted a revised 10-point scale to enhance discriminative capacity. The evaluation prompt template is provided in Appendix~\ref{appendix:critique-prompt}.

\paragraph{Topical Relevance} We evaluated how well generated content aligns with the target research topic, following the approach in \textsc{AutoSurvey}~\cite{wang2024autosurvey}. Our assessment focused on whether the survey maintains consistent focused on the intended subject, avoiding off-topic content. We employed a five-level scoring rubric (Table~\ref{tab:topic_relevance}) that measures increasing degrees of topical coherence. We preserved the original \textsc{AutoSurvey} rubric without modification for comparability with prior work.

\renewcommand{\arraystretch}{1.5}

\begin{table}[ht]
    \small
    \centering
    \begin{tabular}{>{\centering\arraybackslash}m{1.5cm} >{\centering\arraybackslash}m{12cm}}
    \toprule
    \textbf{Score} & \textbf{Description} \\
    \midrule
    1 & The  content is outdated or unrelated to the field it purports to review, offering no alignment with the topic \\
    \hline
    2 & The survey is somewhat on topic but with several digressions; the core subject is evident but not consistently adhered to. \\
    \hline
    3 & The survey is generally on topic, despite a few unrelated details. \\
    \hline
    4 & The survey is mostly on topic and focused; the narrative has a consistent relevance to the core subject with infrequent digressions. \\
    \hline
    5 & The survey is exceptionally focused and entirely on topic; the article is tightly centered on the subject, with every piece of information contributing to a comprehensive understanding of the topic. \\
    \bottomrule
    \end{tabular}
    \vspace{6pt}   
    \caption{Topical Relevance Assessment Rubric.}
    \label{tab:topic_relevance}
\end{table}

\subsubsection{Structural Coherence Assessment}
\label{sec:eval-structure}

We evaluated the structural quality of generated content from both local and global perspectives.

\paragraph{Section-Level Structure} This dimension assesses the internal coherence and logical flow of individual sections and subsections, following the rubric proposed in \textsc{AutoSurvey}. A score of 1 indicates disorganized or incoherent content, while a score of 5 denotes a tightly structured and logically consistent organization with smooth transitions. After evaluation, the score is linearly scaled to a 0–100 range. The full rubric is provided in Table~\ref{table:struct_eval}.

\begin{table*}[ht]
    \small
    \centering
    \begin{tabular}{>{\centering\arraybackslash}m{1.5cm} >{\centering\arraybackslash}m{12cm}}
    \toprule
    \textbf{Score} & \textbf{Description} \\
    \midrule
    1 & The survey lacks logic, with no clear connections between sections, making it difficult to understand the overall framework. \\
    \hline
    2 & The survey has weak logical flow with some content arranged in a disordered or unreasonable manner. \\
    \hline
    3 & The survey has a generally reasonable logical structure, with most content arranged orderly, though some links and transitions could be improved such as repeated subsections.\\
    \hline
    4 & The survey has good logical consistency, with content well arranged and natural transitions, only slightly rigid in a few parts. \\
    \hline
    5 & The survey is tightly structured and logically clear, with all sections and content arranged most reasonably, and transitions between adajecent sections smooth without redundancy. \\
    \bottomrule
    \end{tabular}
    \vspace{6pt}
    \caption{Structural Coherence Evaluation Rubric.}
    \label{table:struct_eval}
\end{table*}

\paragraph{Document-Level Structure}
This dimension evaluates the overall structural coherence, thematic completeness, and scholarly depth of the document structure. We adopted a composite scoring scheme, assigning a score from 0 to 10 for each of the following three criteria: (1) structural coherence and narrative logic, (2) conceptual depth and thematic coverage, and (3) critical thinking and scholarly synthesis. The final score is calculated as the average of these sub-scores and is linearly scaled to a 0–100 range. Detailed prompts and scoring criteria are provided in Appendix~\ref{appendix:outline-prompt}.

Note that \textit{Section-Level Structure} focuses on local coherence between adjacent sections or subsections, while \textit{Document-Level Structure} captures the document-wide organization, conceptual design, and thematic rigor.

\subsubsection{Reference Accuracy Assessment}
\label{sec:exp-reference}

To evaluate the quality of reference usage in generated surveys, we compared the reference papers retrieved by models against those cited by human authors using standard information retrieval (IR) metrics. This evaluation is especially critical in retrieval-augmented generation (RAG) settings, as the quality of retrieved content directly impacts the factual accuracy and trustworthiness of the generated text. Specifically, we employed true positives (TP) and the F1 score ~\cite{goutte2005probabilistic} to quantify the degree of alignment between model-selected references and those curated by human experts.

\paragraph{True Positives (TP)}
Let \( A \) denote the set of references retrievd  by the framework, and \( B \) denote the set of references cited in \textsc{Human Written} surveys. We compute the number of correctly predicted references as:
\[
\text{TP} = |A \cap B|.
\]

This metric reflects the absolute count of overlapping references between the model and the human-written baseline.

\paragraph{F1 Score}
We further compute:

\[
    \text{F1} = \frac{2 \cdot \text{Precision} \cdot \text{Recall}}{\text{Precision} + \text{Recall}},
\]
where 
\[
    \text{Precision} = \frac{|A \cap B|}{|A|}, \quad \text{Recall} = \frac{|A \cap B|}{|B|}.
\]

Precision measures the proportion of model-generated references that are also cited by human authors. Recall quantifies the proportion of human-cited references that the model successfully retrieves. The F1 score provides a harmonic mean of these two metrics, offering an overall measure of citation alignment. A higher F1 score indicates stronger agreement with human citation behavior and thus reflects superior reference retrieval quality within the RAG framework.

\subsection{Human Evaluation by Domain Experts}
\label{sec:human-eval}

To provide a more comprehensive evaluation of content quality beyond automatic metrics, we conducted a human study with domain experts. We randomly sampled 10 research topics and recruited graduate-level students in computer science as annotators. For each topic, annotators were presented with two documents: one generated by our \textsc{SciSage} system and one authored by human researchers. They were instructed to compare the texts across multiple dimensions, including logical coherence, academic tone, paragraph transitions, content completeness, and conciseness, among others.

\section{Experiments}
\label{sec:experiment}

\subsection{Baseline Configurations}
\label{sec:exp-baseline}

To assess the effectiveness of \textsc{SciSage}, we compared it against three representative baselines. All methods were implemented using \textsc{Qwen3-32B}~\cite{qwen3}, and the title of each benchmark paper was used as the input seed for generation. Each baseline was executed using its official codebase with default or recommended configurations. A brief description of each baseline is provided below:

\begin{enumerate}
    \item \textbf{OpenScholar (w/ SciSage)}~\cite{asai2024openscholar}: Since OpenScholar did not support outline generation natively, we incorporated outlines and paragraph-level queries generated by \textsc{SciSage} into its pipeline. The implementation was based on the official repository (\url{https://github.com/AkariAsai/OpenScholar}), and both local and online retrieval mode were enabled.

    \item \textbf{AutoSurvey}~\cite{wang2024autosurvey}: As \textsc{AutoSurvey} lacks support for online retrieval, we use its offline corpus for both retrieval and summarization. Our implementation strictly follows the official codebase (\url{https://github.com/AutoSurveys/AutoSurvey}).

    \item \textbf{LLM~$\times$~MapReduce-V2}~\cite{wang2025llm}: We followed the official implementation from \url{https://github.com/thunlp/LLMxMapReduce}. The paper title was directly used as the input query, and the system employed its built-in online retrieval mechanism to collect relevant content before generation.

\end{enumerate}

Complete hyperparameter settings for each baseline are provided in Appendix~\ref{appendix:baseline_hyperparameters}.

\subsection{Main Result}
\label{sec:exp-main-res}

\subsubsection{Automatic Evaluation Results}
\label{sec:exp-auto-results}

All evaluation results were obtained using \textsc{Qwen3-32B}~\cite{qwen3}. Table~\ref{tab:main_result} reports the automatic evaluation scores for \textsc{SciSage} and three competitive baselines across content quality, structural coherence, reference accuracy.

\textbf{Content Quality.} \textsc{SciSage} achieves the highest score in critical thinking (77.58) while maintaining strong language fluency (85.65), slightly below \textsc{LLM~×~MapReduce-V2} (86.14). It also achieves perfect topical relevance (100). These results suggest that \textsc{SciSage} generally produces higher-quality content.

\textbf{Structural Coherence.} At both the section and document levels, \textsc{SciSage} outperformed all baselines, with the highest document coherence score (80.37). This indicates that \textsc{SciSage} demonstrates superior logical flow and structural organization.

\textbf{Reference Accuracy.} \textsc{SciSage} substantially improves citation accuracy, achieving an F1 score of 0.46 by correctly matching 1,510 references out of 3,844 cited in \textsc{Human Written} papers. In contrast, competing baselines typically retrieve only a single overlapping reference, highlighting their limited capability in accurate citation reproduction.

These evaluation results demonstrate the effectiveness of \textsc{SciSage}. It consistently outperforms baselines across almost all metrics, especially in reference accuracy and document-level coherence.

\begin{table*}[ht]
    \centering
    \begin{tabular}{c|ccc|cc|cc}
    \toprule
    \multirow{2}{*}{\textbf{Method}} & \multicolumn{3}{c|}{\textbf{Content Quality}} & \multicolumn{2}{c|}{\textbf{Structural Coherence}} & \multicolumn{2}{c}{\textbf{Reference}} \\ \cline{2-8}
    & Language & Critical & Relevance & Section & Document & F1 & TP  \\ \hline
    OpenScholar (w/ SciSage) & 68.09 & 53.55 & 99 & - & - & 0.061 & 156 \\
    AutoSurvey & 72.13 & 60.90 & 99 & 85 & 65.33 & 0.14 & 392  \\
    LLM~$\times$~MapReduce-V2 & \textbf{86.14} & 76.93 & \textbf{100} & \textbf{100} & 78.64 & 0.017 & 130  \\
    \hline
    SciSage & 85.65 & \textbf{77.58} & \textbf{100} & \textbf{100} & \textbf{80.37} & \textbf{0.46} & \textbf{1510} \\ 
    \bottomrule
    \end{tabular}
    \vspace{6pt}
    \caption{Metrics of Automatic Evaluation.}
    \label{tab:main_result}
\end{table*}

\subsubsection{Human Evaluaion Results}
\label{sec:human-eval-res}

\begin{figure}[ht]
    \centering
    \includegraphics[width=0.45\linewidth]{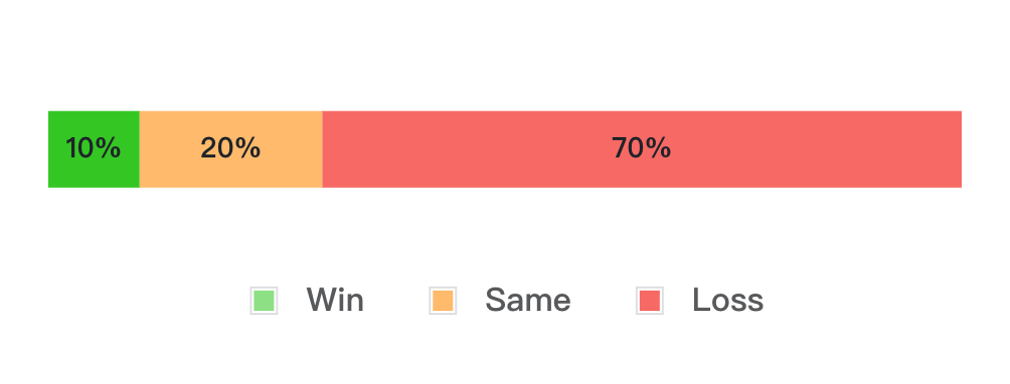}
    \caption{Human evaluation results comparing \textsc{SciSage} with \textsc{Human Written} papers}
    \label{fig:human-evaluation-resul}
\end{figure}

To evaluate the quality of content generated by \textsc{SciSage}, we conducted a human evaluation on a randomly selected set of 10 papers. These papers were assessed by professional evaluators, each holding a Master's degree in Computer Science. The evaluators performed a comprehensive analysis, contrasting the characteristics and identified shortcomings of \textsc{SciSage}'s output against content authored by expert researchers on identical topics. Figure~\ref{fig:human-evaluation-resul} shows the human evaluation results between \textsc{SciSage} and \textsc{Human Written}, with further details provided in Appendix~\ref{appendix:huaman-eval-details}.

\paragraph{Strengths: Broad Coverage and Summarization}
\textsc{SciSage} excels at generating content that is broad in scope and performs as well as or better than human authors on summarization tasks. For example, in areas requiring extensive literature reviews and synthesis, such as the "\textit{Reasoning with Large Language Models, a Survey}", \textsc{SciSage} can effectively summarize and present information. This feature makes it a valuable tool for quickly generating overviews and synthesizing large amounts of information.

\paragraph{Limitations: Depth, Precision, and Stylistic Nuance}
\textsc{SciSage} faces significant challenges in terms of content depth, especially when dealing with complex arguments, subtle details, and scenarios that require rigorous logical coherence or empirical support. It lacks precise and rigorous mathematical expression, which is particularly prominent in fields such as "reinforcement learning and algorithm research" that rely on precise formula descriptions. In terms of language style, \textsc{SciSage} tends to complicate sentence structure, resulting in less clear and concise writing, for example, by using vague terms such as "mitigation techniques" instead of precise academic vocabulary. Similar to existing generative models or frameworks, \textsc{SciSage}'s generation also suffers from lengthy text and lacks integrated visual elements such as detailed formulas/charts, which are key carriers for conveying complex information in academic communication.

\paragraph{Conclusion: SciSage's Capabilities and Limitations in Academic Content Generation}
\textsc{SciSage} performs well in literature review and information integration, and can efficiently generate academic content with wide coverage, even surpassing the level of professional human authors. However, it still lacks analytical depth, mathematical expression accuracy, and academic language style, especially in fields that require complex logical reasoning, precise formulas, or rigorous terminology.

\section{Ablation Study}
\label{sec:ablation study}

\subsection{Structural Impact of Query Understanding}
\label{sec:ablation-structure}

We conduct an ablation study to investigate the structural benefits introduced by incorporating \textbf{Query Understanding} (Q.U.) in our framework. Specifically, we compare the following two experimental settings:

\begin{itemize}
    \item \textbf{Experiment A (w/ Q.U.):} The complete \textsc{SciSage} pipeline, where the system first performs query understanding before generating the document structure.
    \item \textbf{Experiment B (w/o Q.U.):} A simplified pipeline that omits the query understanding step and directly proceeds to structure generation.
\end{itemize}

We evaluated the structural quality of the generated outlines along three dimensions: structural coherence, topical coverage, and critical analysis, as defined in Appendix~\ref{appendix:outline-prompt}.

As shown in Table~\ref{tab:query_understanding_ablation}, incorporating query understanding leads to consistent improvements in both overall and aspect-level evaluation. The average and maximum document-level scores increase from 8.04 to 8.16 and from 9.00 to 9.33, respectively. Aspect-wise, improvements are observed in structure (8.74 vs. 8.64), coverage (8.32 vs. 8.20), and analysis (7.40 vs. 7.29). These results suggest that query understanding enhances the \textsc{SciSage}’s ability to generate outlines that are more coherent, comprehensive, and analytically robust. (Full evaluation details and examples are provided in our project repository.)

\begin{table*}[ht]
    \centering
    \begin{tabular}{c|ccc|ccc}
    \toprule
    \multirow{2}{*}{\textbf{Method}} & \multicolumn{3}{c|}{\textbf{Document Level Structure}} & \multicolumn{3}{c}{\textbf{Structure Score Details}} \\ \cline{2-7}
    & \textbf{Avg} & \textbf{Max} & \textbf{Min} & \textbf{Structure} & \textbf{Coverage} & \textbf{Analysis} \\ 
    \hline
    w/o Q.U. & 8.04 & 9.00 & \textbf{6.33} & 8.64 & 8.20 & 7.29 \\
    w/ Q.U.  & \textbf{8.16} & \textbf{9.33} & 6.00 & \textbf{8.74} & \textbf{8.32} & \textbf{7.40} \\
    \bottomrule
    \end{tabular}
    \vspace{6pt}
    \caption{Comparison of \textsc{SciSage} with and without Reflection.}
    \label{tab:query_understanding_ablation}
\end{table*}

\subsection{Contribution of the Reflection}
\label{sec:ablation-refelec}

To assess the impact of iterative hierarchical reflection, we conducted an ablation study by disabling the reflection component in \textsc{SciSage}. Table~\ref{tab:reflection-result} presents a comparison between the full system and its ablated variant.

Results show that reflection leads to sustained improvements in all dimensions assessed. Specifically, content quality improved significantly: \textbf{Language} scores increased from 82.28 to 85.60, and \textbf{Critical} scores significantly increased from 69.70 to 77.93. Structural coherence also benefited from reflection, with \textbf{Document}-level structure scores improving from 71.25 to 81.48. These findings suggest that repeated reflection enables \textsc{SciSage} to better revise and organize its generated content, resulting in more fluent, thoughtful, and well-structured content.

\begin{table*}[ht]
    \centering
    \begin{tabular}{c|ccc|cc}
    \toprule
    \multirow{2}{*}{\textbf{Method}} & \multicolumn{3}{c|}{\textbf{Content Quality}} & \multicolumn{2}{c}{\textbf{Structural Coherence}} \\ \cline{2-6}
    & \textbf{Language} & \textbf{Critical} & \textbf{Relevance} & \textbf{Section} & \textbf{Document} \\ 
    \hline
    SciSage (w/o Reflection) & 82.28 & 69.70 & \textbf{100.00} & 99.00 & 71.25 \\
    SciSage (w/ Reflection)  & \textbf{85.60} & \textbf{77.93} & \textbf{100.00} & \textbf{100.00} & \textbf{81.48} \\
    \bottomrule
    \end{tabular}
    \vspace{6pt}
    \caption{Comparison of \textsc{SciSage} with and without Reflection.}
    \label{tab:reflection-result}
\end{table*}

\section{Limitations}

Our study has several limitations that should be acknowledged:

\begin{itemize}
\item \textbf{Language Restriction}: The current evaluation is limited to English-language queries and documents. The effectiveness of our approach for other languages (e.g., Chinese) remains untested and may require additional language-specific adaptations.

\item \textbf{Domain Specificity}: While we demonstrate strong performance in academic paper retrieval, the generalizability of our method to broader search scenarios (e.g., web search or enterprise document retrieval) requires further validation.

\item \textbf{Model Dependence}: All reported results are based on the \textsc{Qwen3-32B}\cite{qwen3}. The performance characteristics may vary when implemented with other foundation models, and comprehensive cross-model evaluation would be needed to establish broader applicability.

\item \textbf{Metric Saturation}: Several systems, including \textsc{SciSage}, \textsc{LLM~×~MapReduce-V2}, and \textsc{AutoSurvey}, achieved near-perfect scores in both \textit{Topical Relevance} and \textit{Section Coherence}. This saturation suggests that these metrics are becoming less effective in distinguishing between modern LLM-based generation systems, as they typically produce well-structured and topically relevant content. Future evaluations may require more fine-grained metrics to capture subtle differences in reasoning and factual consistency.

\end{itemize}

\section{Conclusion}
In this work, we present SciSage, a novel multi-agent framework that addresses long-standing limitations in automated scientific survey generation—specifically issues of structural coherence, content depth, and citation reliability. Guided by a \textit{reflect-when-you-write} paradigm, SciSage coordinates six specialized agents across a dynamic workflow, with the Reflector Agent playing a central role in iteratively critiquing and refining outputs at the outline, section, and document levels. This reflection-driven architecture emulates expert authoring behavior and ensures end-to-end consistency and factual accuracy throughout the generation pipeline.
\textsc{SciSage} significantly improves structural coherence and citation accuracy over existing methods. To rigorously evaluate system performance, we introduce \textbf{SurveyScope} benchmark, curated for recency and scholarly impact, provides a robust testbed for evaluating survey-generation systems. Empirical results confirm \textsc{SciSage}’s superiority: it achieves an 80.37 document-coherence score (vs. 78.64 for LLM×MapReduce-V2) and 46\% citation F1, outperforming all baselines. 
While \textsc{SciSage} still trails human-authored surveys in analytical depth (30\% win rate), it demonstrates clear advantages on relatively straightforward topics and offers substantial reductions in drafting time, highlighting its practical utility.

\bibliographystyle{unsrt}  
\bibliography{references}  
\newpage

\appendix

\section{Prompt Template}
\label{appendix:prompt template}

\subsection{Prompt for Query Understanding}
\label{appendix:prompt for query understanding}

\begin{tcolorbox}[
  colback=white!10!white,
  colframe=black!75!black,
  fonttitle=\bfseries,
  breakable,
  sharp corners=south,
  enhanced,
  width=\textwidth,
  title=\textbf{Prompt for Query Intent Chassification},
]
  You are an expert in classifying user queries for academic research purposes.
  Your task is to analyze the given user query and extract the following information:

  \begin{enumerate}
    \item \textbf{Research Domain}: Identify the broad academic field the query falls into.
    Examples: Computer Science, Medicine, Physics, Sociology, History, Linguistics.
    Be as specific as reasonably possible (e.g., "Machine Learning" if clearly indicated within Computer Science, otherwise "Computer Science").

    \item \textbf{Query Type}: Determine the type of information or paper the user is likely seeking.
    You MUST choose one of the following predefined types:
    \texttt{survey, method, application, analysis, position, theory, benchmark, dataset, OTHER}.
    If none of the specific types fit well, use \texttt{OTHER}.

    \item \textbf{Research Topic}: Pinpoint the specific subject, concept, or entities at the core of the query.
    This should be a concise phrase representing the main focus. For example, if the query is "latest advancements in using LLMs for code generation", the topic could be "LLMs for code generation".
  \end{enumerate}
\end{tcolorbox}

\noindent\rule{\linewidth}{0.5pt}

\begin{tcolorbox}[
  colback=white!10!white,
  colframe=black!75!black,
  title=\textbf{Prompt for Query Rewriting},
  fonttitle=\bfseries,
  breakable,
  sharp corners=south,
  enhanced,
  width=\textwidth
]
  You are a query rewriting expert.
  Your task is to evaluate a given query and determine if it requires rewriting by checking for:

  \begin{enumerate}
    \item Semantic clarity issues
    \item Ambiguity
    \item Contextual fit for search/research scenarios
    \item Overly complex or verbose phrasing
  \end{enumerate}

  If rewriting is needed, create a revised version that:
  \begin{itemize}
    \item Maintains the original semantic meaning
    \item Is more precise and concise
    \item Is better suited for search/research purposes
  \end{itemize}
\end{tcolorbox}

\subsection{Prompt for Benchmark Topic Classification}
\label{appendix:Prompt for Topic Classification}

\begin{tcolorbox}[
  colback=white!10!white,
  colframe=black!75!black,
  fonttitle=\bfseries,
  breakable,
  sharp corners=south,
  enhanced,
  width=\textwidth
    ]
    You are an expert in computer science research. Based on the following paper title, please complete the two tasks below:
    
    \begin{enumerate}
          \item Extract the main research topic of the paper (expressed as a concise phrase, such as: \textit{Robustness in NLP Models}, \textit{Multimodal Learning}, \textit{LLM Safety}, etc.).
          \item Assign the extracted topic to one of the following high-level categories:
    \end{enumerate}
        
    \textbf{Category List:}
    \begin{itemize}
      \item[1.] NLP  
      \item[2.] LLMs (General)  
      \item[3.] LLMs Safety  
      \item[4.] LLMs Efficiency  
      \item[5.] Dialogue Systems  
      \item[6.] Multimodal  
      \item[7.] Medical / Biomedical  
      \item[8.] Finance / Domain-specific  
      \item[9.] Robotics  
      \item[10.] Benchmarking / Evaluation  
      \item[11.] Other  
    \end{itemize}
    
    \textbf{Paper Title:} \textcolor{blue}{\texttt{\{title\}}}
    
    \vspace{0.5em}
    \textbf{Please return the result in the following format:}
    \begin{verbatim}
    Research Topic: [your topic]
    Category: [your chosen category]
    \end{verbatim}
\end{tcolorbox}

\subsection{Prompt for Benchmark Topic Completion}
\label{appendix:Prompt for Topic Completion}

\begin{tcolorbox}[
      colback=white!10!white,
      colframe=black!75!black,
      fonttitle=\bfseries,
      breakable,
      sharp corners=south
    ]
    
    You are an expert in computer science research. Now I want to gain a comprehensive overview of the current research hotspots across the field. Below is a list of topics I have already identified:\\
    
    Topic List:\textbf{ \{topic\_list\}}\\
    
    Please analyze the list and suggest any important research directions or topics that are currently missing, in order to make the coverage more complete and representative of the field.
\end{tcolorbox}

\subsection{Pormpt for Evaluation Language Fluency Score}
\label{appendix:language-prompt}

\begin{tcolorbox}[
  colback=white,
  colframe=black!75,
  fonttitle=\bfseries,
  breakable,
  sharp corners=south,
    ]
    
    \noindent\textbf{[Task]}\\
    Rigorously evaluate the quality of an academic survey on the topic of \texttt{[TOPIC]} by scoring three dimensions on a 0--10 scale. The final score is the arithmetic mean of the three individual scores.
    
    \vspace{0.5em}
    \noindent\textbf{[Evaluation Criteria]}\\
    Assign scores for each dimension based on the highest academic standards described below. The final score is calculated as the average of the three:
    
    \begin{enumerate}
      \item \textbf{Academic Formality (10 points)}\\
      Demonstrates \textit{flawless} academic rigor. Uses precise terminology consistently, avoids colloquial language entirely, and maintains a scholarly tone throughout. Sentence structures are sophisticated and intentionally crafted to support analytical depth. \textbf{Even a single instance of informal phrasing or vague terminology disqualifies a perfect score}.
      
      \item \textbf{Clarity \& Readability (10 points)}\\
      Writing is \textit{exceptionally} clear, concise, and unambiguous. Sentences are logically structured with seamless transitions. The argument progresses smoothly with no unnecessary complexity. \textbf{Any ambiguity or minor inefficiency reduces the score}.
      
      \item \textbf{Redundancy (10 points)}\\
      \textbf{Uniqueness}: Every sentence should contribute new value. Repetition is only acceptable for structural clarity, such as reinforcing terminology or aiding transitions.\\
      \textbf{Efficiency}: Arguments must be logically coherent and free from unnecessary repetition. Redundant rephrasing of the same point without adding new insight leads to point deductions.
    \end{enumerate}
    
    \vspace{0.5em}
    \noindent\textbf{[Topic]}\\
    \textcolor{blue}{\texttt{[TOPIC]}}
    
    \vspace{0.5em}
    \noindent\textbf{[Section]}\\
    \textcolor{blue}{\texttt{[SECTION]}}
    
    \vspace{0.5em}
    \noindent\textbf{[Output Format]}\\
    \texttt{Rationale:}\\
    \textless Provide a detailed justification for the score. Discuss each dimension individually, highlighting specific strengths and weaknesses (e.g., academic tone consistency, clarity of sentence structure, or presence of redundancy).\textgreater
    
    \vspace{0.3em}
    \texttt{Final Score:}\\
    \textless SCORE\textgreater(\textit{X+Y+Z}/3 = \textbf{Final})\textless/SCORE\textgreater\\
    \textit{Example:} <SCORE>(2.5+7+5.1)/3=4.87</SCORE> \\
    \textit{Use up to two decimal places. Do not include any text outside the SCORE tags.}
    
\end{tcolorbox}

\subsection{Pormpt for Evaluation Critical Thinking Score}
\label{appendix:critique-prompt}

\begin{tcolorbox}[
  colback=white,
  colframe=black!75,
  fonttitle=\bfseries,
  breakable,
  sharp corners=south,
    ]
    \noindent\textbf{[Task]}\\
    Rigorously evaluate the quality of an academic survey on the topic of \texttt{[TOPIC]} by scoring three dimensions (each on a 0--10 scale) and computing the average as the final score.
    
    \vspace{0.75em}
    \noindent\textbf{[Evaluation Criteria]}\\
    The final score is the average of the individual scores from the following three dimensions. Please evaluate each dimension rigorously based on the highest scholarly standards.
    
    \begin{enumerate}
      \item \textbf{Critical Analysis (10 points)}\\
      Offers a deep and incisive critique of methodologies, results, and underlying assumptions. Clearly identifies significant gaps, weaknesses, and areas for improvement. Challenges assumptions with well-supported arguments and proposes concrete alternatives.
    
      \item \textbf{Original Insights (10 points)}\\
      Proposes novel, well-supported interpretations or frameworks based on the reviewed literature. Demonstrates strong subject-matter understanding and contributes genuinely original perspectives. Insights are well-integrated with existing research, challenging conventional views or offering new directions.
    
      \item \textbf{Future Directions (10 points)}\\
      Clearly articulates promising research directions with strong justification. Suggestions are concrete, actionable, and closely tied to gaps identified in the literature. Demonstrates foresight by proposing innovative approaches or methodologies.
    \end{enumerate}
    
    \vspace{0.75em}
    \noindent\textbf{[Topic]}\\
    \textcolor{blue}{\texttt{[TOPIC]}}
    
    \vspace{0.75em}
    \noindent\textbf{[Section]}\\
    \textcolor{blue}{\texttt{[SECTION]}}
    
    \vspace{0.75em}
    \noindent\textbf{[Output Format]}\\
    \texttt{Rationale:}\\
    \textless Provide a detailed justification for the score. Address each of the three dimensions step by step, highlighting specific strengths and weaknesses, such as the depth of critique, the originality of insights, or the clarity of proposed future directions. \textgreater
    
    \vspace{0.5em}
    \texttt{Final Score:}\\
    \textless SCORE\textgreater(\textit{X+Y+Z}/3 = \textbf{Final})\textless/SCORE\textgreater\\
    \textit{Example:} \texttt{<SCORE>(2.5+7+5.1)/3=4.87</SCORE>} \\
    \textit{Use two decimal places; do not include any other text outside the SCORE tag.}
\end{tcolorbox}

\subsection{Prompt for Evaluation Document Outline}
\label{appendix:outline-prompt}

\begin{tcolorbox}[
  colback=white,
  colframe=black!75,
  fonttitle=\bfseries,
  breakable,
  sharp corners=south,
    ]
    \noindent\textbf{[Task]}\\
    Rigorously evaluate the quality of an academic survey \textbf{outline} on the topic of \texttt{[TOPIC]} by scoring three dimensions (each on a 0--10 scale) and computing the average as the final score.
    
    \vspace{0.75em}
    \noindent\textbf{[Evaluation Criteria]}\\
    Evaluate each dimension on a strict 0--10 scale, based on the following high-precision standards. The final score is the average of the three dimension scores.
    
    \begin{enumerate}
      \item \textbf{Structural Coherence \& Narrative Logic (10 points)}\\
      \textbf{Ideal Standard}: The outline presents a well-structured, logically flowing framework. Sections and subsections are clearly organized, transitions are smooth, and the narrative progression is coherent.\\
      \textbf{Scoring Guidance}: Deduct points for imbalanced section lengths, disjointed transitions, or subsections that interrupt narrative clarity. A perfect score (10) requires no observable flaws.
    
      \item \textbf{Conceptual Depth \& Thematic Coverage (10 points)}\\
      \textbf{Ideal Standard}: The outline captures key themes, concepts, and subfields comprehensively and insightfully. There is a balance of breadth and depth, with core debates and historical development of the field clearly reflected.\\
      \textbf{Scoring Guidance}: Deduct points for missing major themes, excessive focus on niche areas, or shallow treatment of foundational concepts.
    
      \item \textbf{Critical Thinking \& Scholarly Synthesis (10 points)}\\
      \textbf{Ideal Standard}: The outline integrates perspectives critically, addressing contradictions, methodological tensions, and open research questions. It synthesizes viewpoints into a coherent scholarly vision.\\
      \textbf{Scoring Guidance}: Deduct points for lack of critical analysis, overlooking disagreements or critiques, or failing to propose unresolved questions.
    \end{enumerate}
    
    \vspace{0.75em}
    \noindent\textbf{[Topic]}\\
    \textcolor{blue}{\texttt{[TOPIC]}}
    
    \vspace{0.75em}
    \noindent\textbf{[Skeleton]}\\
    \textcolor{blue}{\texttt{[OUTLINE]}}
    
    \vspace{0.75em}
    \noindent\textbf{[Output Format]}\\
    \texttt{Rationale:}\\
    \textless Provide a detailed reason for the score, considering each dimension step by step. Highlight specific strengths and weaknesses, such as structural imbalances, thematic omissions, or weak analytical synthesis. Then provide the final scores for each dimension. \textgreater
    
    \vspace{0.5em}
    \texttt{- Structure: <X/10>}\\
    \texttt{- Coverage: <Y/10>}\\
    \texttt{- Critical Analysis: <Z/10>}\\
    
    \vspace{0.5em}
    \texttt{Final Score:}\\
    \textless SCORE\textgreater(\textit{X+Y+Z}/3 = \textbf{...})\textless/SCORE\textgreater\\
    \textit{Example:} \texttt{<SCORE>(2.5+7+5.1)/3=4.87</SCORE>} \\
    \textit{Use two decimal places; do not include any other text outside the SCORE tag.}
\end{tcolorbox}

\section{Benchmark}

\subsection{Category Comparision}
\label{appendix:benchmark-compare-category}

\begin{figure}[ht]
    \centering
     \includegraphics[width=0.6\linewidth]{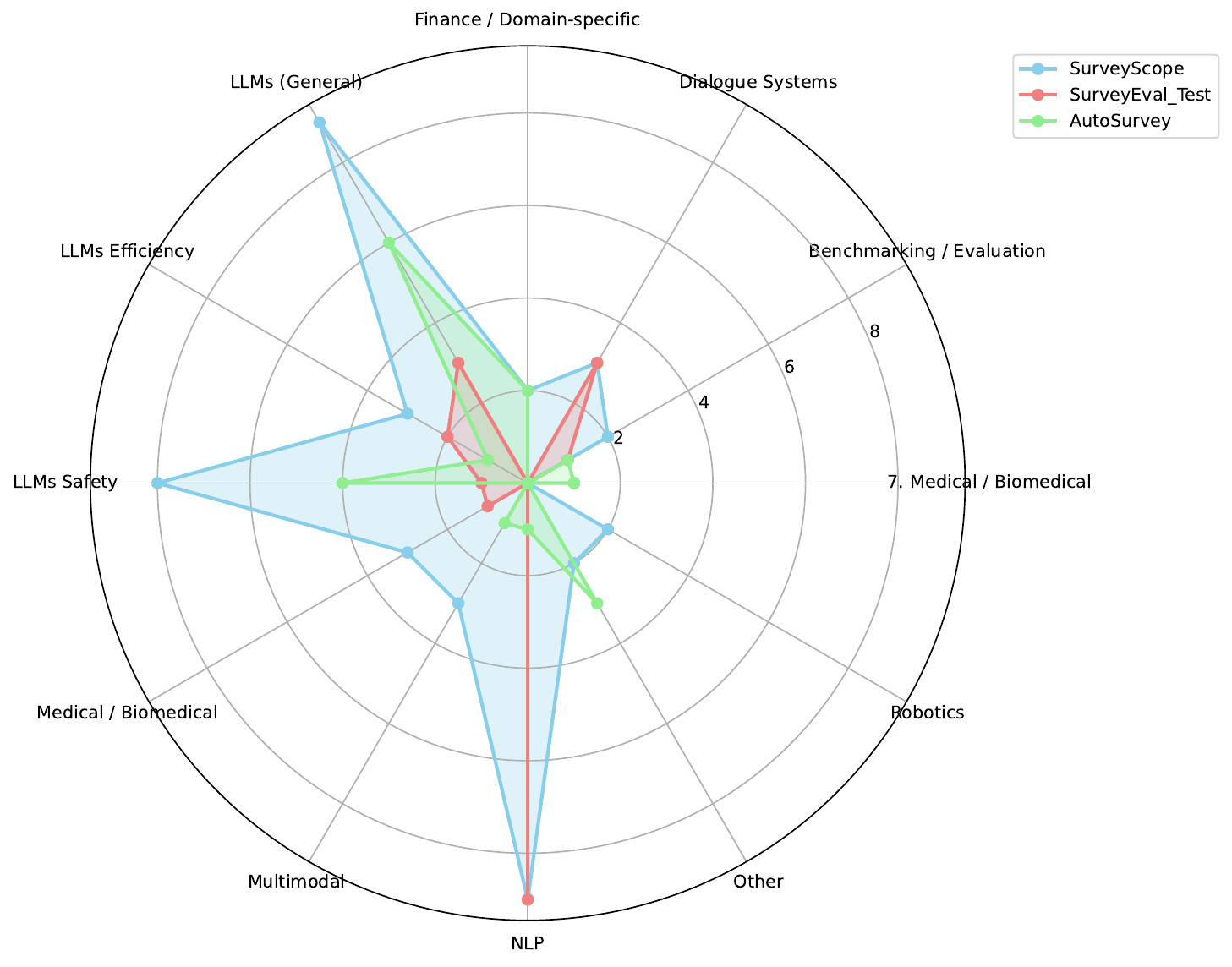}
    \caption{Radar chart illustrating topic distribution across \textbf{SurveyScope}, \textsc{SurveyEval\_Test}, and \textit{AutoSurvey}. \textbf{SurveyScope} exhibits broader and more balanced domain coverage.}
\end{figure}

\subsection{Publication Year Comparision}
\label{appendix:benchmark-compare-recenty}

\begin{figure}[ht]
    \centering
     \includegraphics[width=0.6\linewidth]{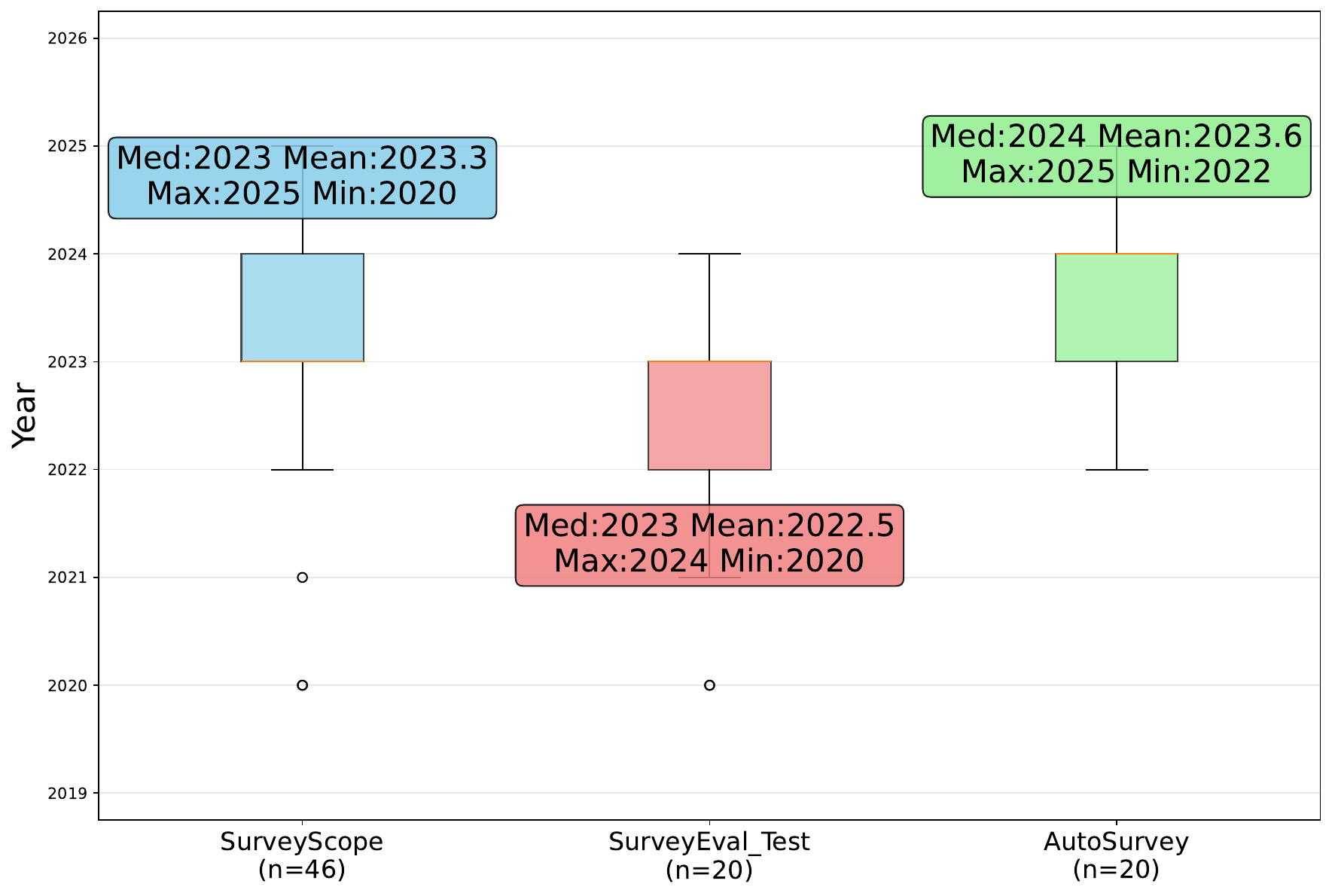}
    \caption{Boxplot showing publication year distributions across benchmarks. \textbf{SurveyScope} emphasizes more recent works, reflecting rapid developments in the field.}
\end{figure}

\subsection{Citation Comparision}
\label{appendix:benchmark-compare-citation}

\begin{figure}[htbp]
    \centering
    \includegraphics[width=0.6\linewidth]{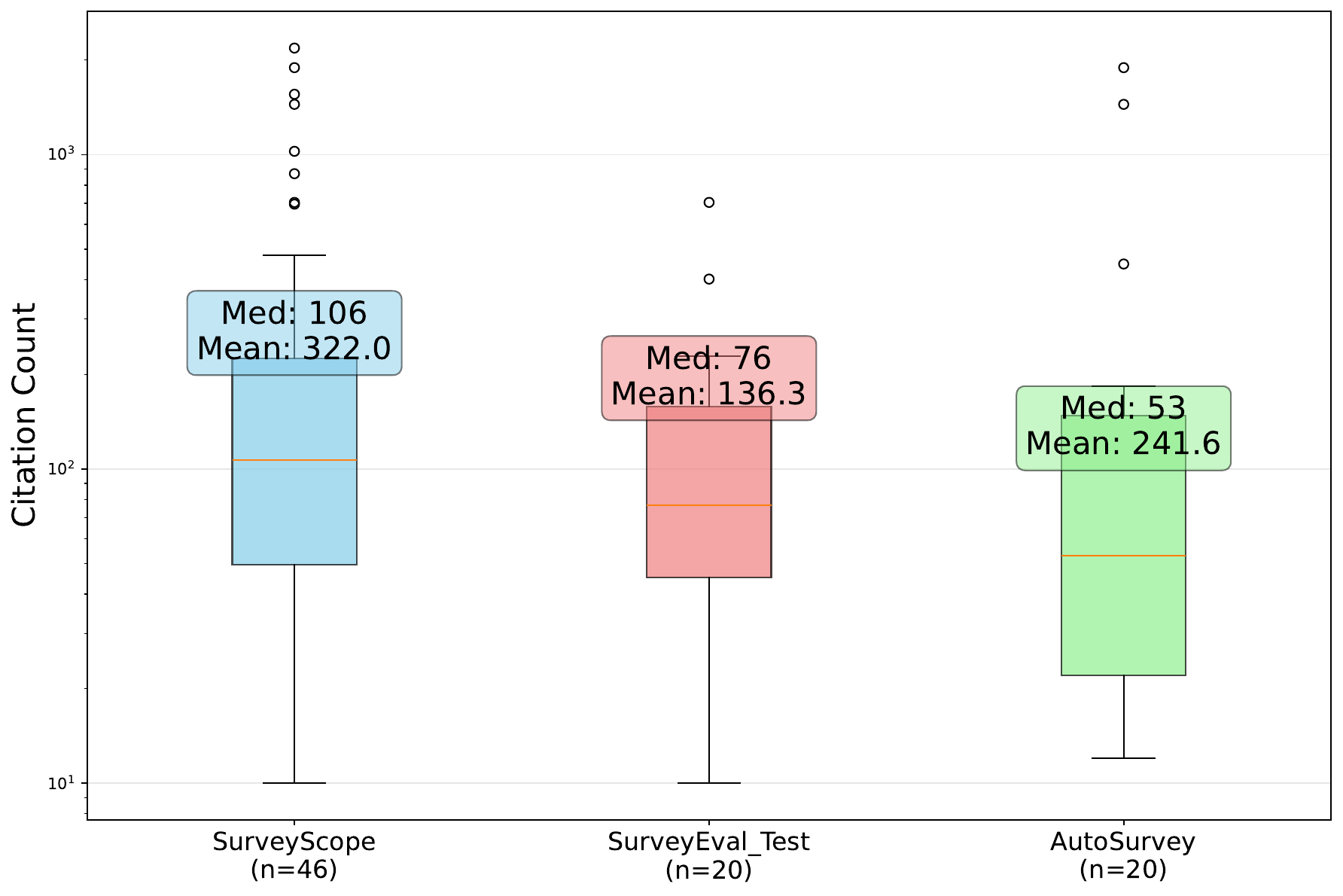}
    \caption{Boxplot of citation counts across benchmarks. Papers in \textbf{SurveyScope} show higher citation impact than those in \textit{SurveyEval\_Test} and \textit{AutoSurvey}.}

\end{figure}

\subsection{Benchmarks Comparision}
\label{appendix:benchmark-compare-details}

\begin{figure}[htbp]
    \centering
    \includegraphics[width=0.9\linewidth]{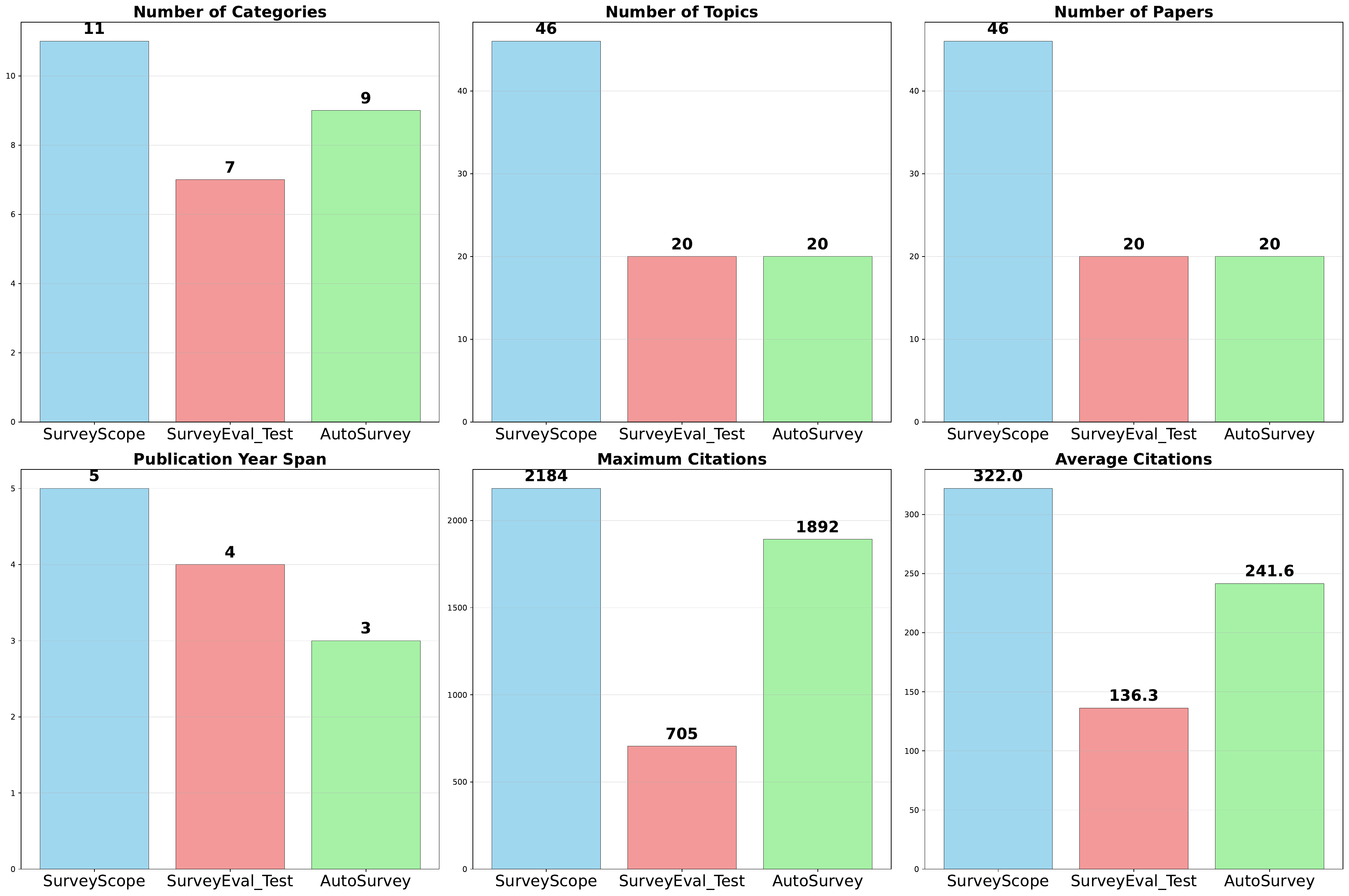}
    \caption{Comparsion of differents benchmarks}

\end{figure}

\section{Experiment Settings}
\label{appendix:baseline_hyperparameters}

\begin{tcolorbox}[
      colback=white!10!white,
      colframe=black!75!black,
      title=\textbf{SciSage Experiment Settings},
      fonttitle=\bfseries,
      breakable,
      sharp corners=south
    ]
    \textbf{search\_url}: https://serper.dev, https://api.openalex.org/works \hfill \textit {Retrival Url} \\
    \textbf{outline\_max\_reflections}: 2 \hfill \textit{Number of structural reflection iterations.}\\
    \textbf{outline\_max\_sections}: 6 \hfill \textit{Maximum number of outline sections.}\\
    \textbf{outline\_min\_depth}: 2 \hfill \textit{Minimum depth of outline hierarchy.}\\
    \textbf{outline\_generate\_models}: [Qwen3-14B,Qwen3-32B,Llama3-70B] \hfill \textit{Generation models for outline.}\\
    \textbf{section\_writer\_model}: Qwen3-32B \hfill \textit{Model used for paragraph generation.}\\
    \textbf{do\_section\_reflection}: True \hfill \textit{Enable paragraph-level reflection.}\\
    \textbf{section\_reflection\_model}: Qwen3-32B \hfill \textit{Model used for section reflection.}\\
    \textbf{section\_reflection\_max\_turns}: 2 \hfill \textit{Maximum paragraph reflection rounds.}\\
    \textbf{do\_global\_reflection}: True \hfill \textit{Enable global-level reflection.}\\
    \textbf{global\_reflection\_max\_turns}: 2 \hfill \textit{Maximum global reflection rounds.}\\
    \textbf{global\_abstract\_conclusion\_max\_turns}: 1 \hfill \textit{Reflection rounds for abstract and conclusion.}

\end{tcolorbox}

\begin{tcolorbox}[
      colback=white!10!white,
      colframe=black!75!black,
      title=\textbf{AutoSurvey Experiment Settings},
      fonttitle=\bfseries,
      breakable,
      sharp corners=south
    ]
    \textbf{section\_num}: 7 \hfill \textit{(Number of sections)}\\
    \textbf{subsection\_len}: 700 \hfill \textit{(Words per subsection)}\\
    \textbf{rag\_num}: 60 \hfill \textit{(Number of local RAG retrievals)}\\
    \textbf{outline\_reference\_num}: 1500 \hfill \textit{(References used for outline generation)}\\
    \textbf{embedding\_model}: nomic-embed-text-v1 \hfill \textit{(Embedding model for retrieval)}
\end{tcolorbox}

\vspace{0.5em}
\noindent\rule{\linewidth}{0.5pt}
\vspace{0.5em}

\begin{tcolorbox}[
      colback=white!10!white,
      colframe=black!75!black,
      title=\textbf{LLM~$\times$~MapReduce-V2 Experiment Settings},
      fonttitle=\bfseries,
      breakable,
      sharp corners=south
    ]
    \textbf{block\_count}: 0 \hfill \textit{Number of document blocks.}\\
    \textbf{conv\_layer}: 6 \hfill \textit{Convolution layer count.}\\
    \textbf{conv\_kernel\_width}: 3 \hfill \textit{Convolution kernel width.}\\
    \textbf{conv\_result\_num}: 10 \hfill \textit{Number of results retained.}\\
    \textbf{top\_k}: 6 \hfill \textit{Top-k results for final selection.}\\
    \textbf{search\_url}: https://serper.dev \hfill \textit {Retrival Url}
\end{tcolorbox}

\vspace{0.5em}
\noindent\rule{\linewidth}{0.5pt}
\vspace{0.5em}

\begin{tcolorbox}[
      colback=white!10!white,
      colframe=black!75!black,
      title=\textbf{OpenScholar Experiment Settings},
      fonttitle=\bfseries,
      breakable,
      sharp corners=south
      ]
        \textbf{use\_contexts}: True \hfill \textit{Use retrieved context for generation.}\\
        \textbf{top\_n}: 5 \hfill \textit{Maximum number of documents used per section.}\\
        \textbf{ranking\_ce}: True \hfill \textit{Enable re-ranking with cross-encoder.}\\
        \textbf{min\_citation}: 5 \hfill \textit{Minimum citation count for reference papers.}\\
        \textbf{norm\_cite}: True \hfill \textit{Normalize citation counts.}\\
        \textbf{ss\_retriever}: True \hfill \textit{Enable Semantic Scholar online retrieval.}\\
        \textbf{use\_feedback}: True \hfill \textit{Enable feedback for iterative refinement.}\\
        \textbf{new\_feedback\_docs}: 2 \hfill \textit{Documents retrieved after feedback.}\\
        \textbf{feedback\_num}: 4 \hfill \textit{Number of feedback items used.}
\end{tcolorbox}

\section{Experiment Result}

\subsection{Human Evaluation Details}
\label{appendix:huaman-eval-details}

\begin{table*}[htbp]
    \centering
    \small
    \resizebox{\textwidth}{!}{
    \begin{tabular}{|>{\centering\arraybackslash}p{4.2cm}|
                >{\centering\arraybackslash}p{2.5cm}|
                >{\centering\arraybackslash}p{9cm}|}
    \hline
    \textbf{Paper Title} & \textbf{Evaluation Result} & \textbf{Human Analysis} \\
    \hline
    Measure and Improve Robustness in NLP Models: A Survey & Human is better & Human version defines robustness clearly, has better structure and logic; LLM version has awkward phrasing and lacks coherence. \\
    \hline
    A Survey on Explainability in Machine Reading Comprehension & Human is better & Human version uses structured benchmarks and visuals effectively; LLM version lacks clarity and has poor section design. \\
    \hline
    Efficient Methods for Natural Language Processing: A Survey & \textcolor{orange}{\textbf{Same}} & Both cover NLP efficiency; human is clear, LLM offers broader metrics. \\
    \hline
    The Decades Progress on Code-Switching Research in NLP & Human is better & Human version aligns better with survey goals using empirical analysis; LLM fails to capture research trend focus. \\
    \hline
    A Survey of Large Language Models in Medicine & Human is better & Human version is structured around medical use cases; LLM version is disjointed and overly focused on technical background. \\
    \hline
    A Survey of Controllable Text Generation & Human is better & Human version is intuitive and organized by model stages; LLM version is messy and lacks strategy-method separation. \\
    \hline
    A Survey on Detection of LLMs-Generated Content & \textcolor{orange}{\textbf{Same}} & LLM version is well-structured and easy to follow; human version introduces more novel and timely perspectives. \\
    \hline
    Neural Entity Linking: A Survey of Models Based on Deep Learning & Human is better & Human version follows processing pipeline; LLM version has incoherent topic grouping and surface-level analysis. \\
    \hline
    Reasoning with Large Language Models, a Survey & \textcolor{blue}{\textbf{SciSage is better}} & Human version is CoT-focused but narrow; LLM version covers broader reasoning aspects despite typical stylistic flaws. \\
    \hline
    A Survey on LLM Security and Privacy & Human is better & Human version has unique structure (good/bad/ugly); LLM lacks depth and is disorganized. \\
    \hline
    \end{tabular}
    }
    \caption{Human evaluation details between \textsc{SciSage} and \textsc{Human Written}}
    \label{tab:huamn-eval-details}

\end{table*}
    
\section{Mindmap Example}
\label{appendix:mindmap-example}

\begin{figure}[ht]
    \centering
     \includegraphics[width=0.8\linewidth]{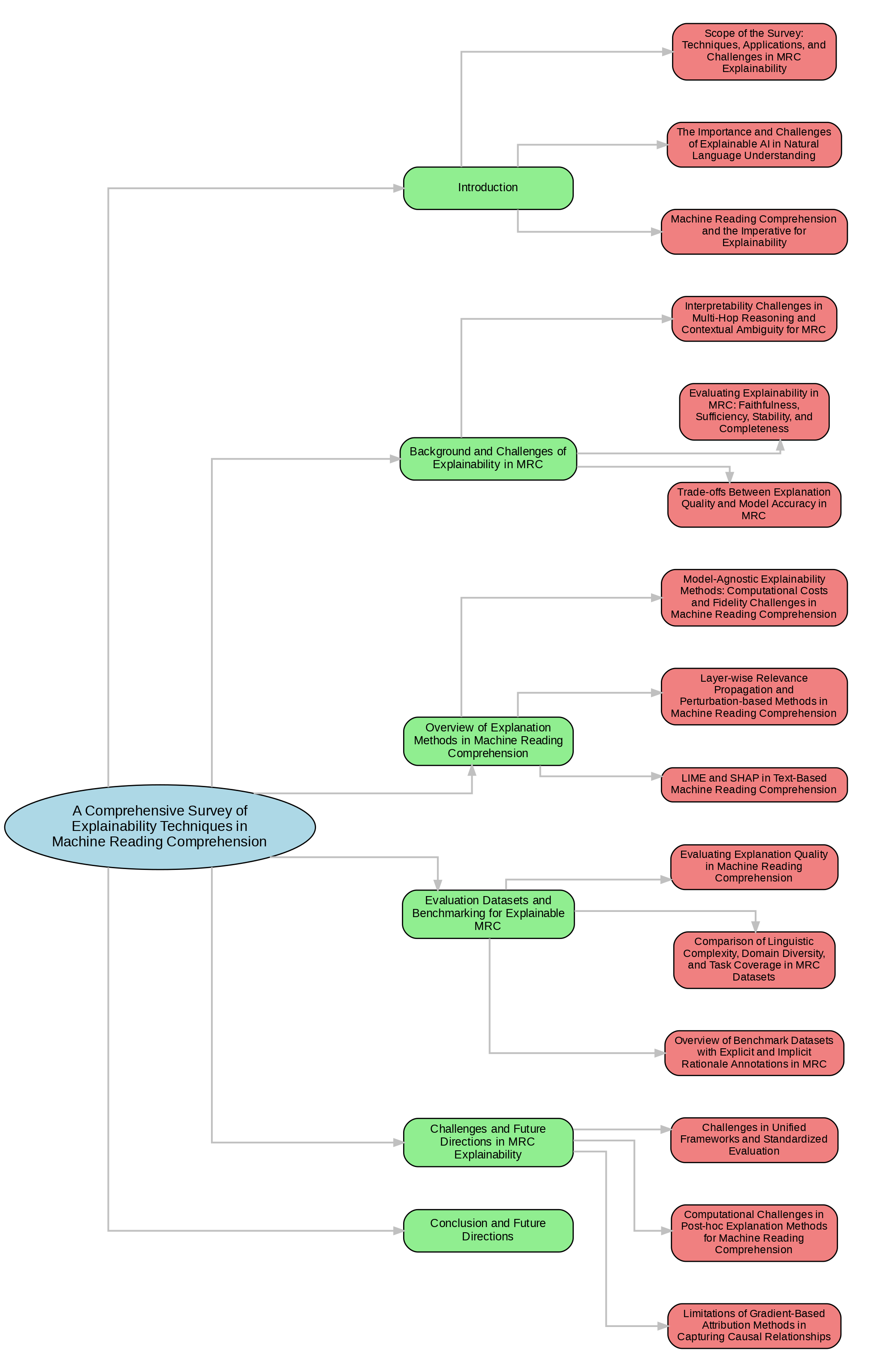}
    \caption{Example of generated outline}
\end{figure}

\end{document}